\documentclass[12pt]{article}
\usepackage{soul}
\usepackage{amssymb}
\usepackage{lineno}
\usepackage{graphicx}
\usepackage{amsmath}
\usepackage{epstopdf}
\usepackage{subfigure}
\usepackage{color}
\usepackage{url}
\usepackage{fancybox}
\usepackage{soul}
\usepackage{cleveref}
\usepackage{natbib}
\bibpunct{(}{)}{;}{a}{,}{,}

\begin{document}
\graphicspath{ {./eps/} }
\setcounter{page}{1}
\pagenumbering{arabic}
\linenumbers
\crefname{figure}{Figure}{Figures}
\newtheorem{mytheorem}{Theorem}
\newtheorem{mylemma}{Lemma}
\newcommand{\argmax}{\mathop{\rm argmax}\limits}
\newcommand{\argmin}{\mathop{\rm argmin}\limits}

\newcommand{\hilight}[1]{{#1}}

\newcommand{\unorm}[1]{\|#1\|}
\newcommand{\unorms}[1]{\unorm{#1}^2}
\newcommand{\calX}{{\mathcal{X}}}
\newcommand{\calY}{{\mathcal{Y}}}
\newcommand{\boldtheta}{{\boldsymbol{\theta}}}
\newcommand{\boldthetaP}{{\boldsymbol{\theta}}^{P}}
\newcommand{\boldthetaQ}{{\boldsymbol{\theta}}^{Q}}
\newcommand{\factorp}{{\phi}^P}
\newcommand{\factorq}{{\phi}^Q}
\newcommand{\boldalpha}{{\boldsymbol{\alpha}}}
\newcommand{\boldHh}{{\widehat{\boldH}}}
\newcommand{\boldH}{{\boldsymbol{H}}}
\newcommand{\boldA}{{\boldsymbol{A}}}
\newcommand{\boldS}{{\boldsymbol{S}}}
\newcommand{\boldK}{{\boldsymbol{K}}}
\newcommand{\boldJ}{{\boldsymbol{J}}}
\newcommand{\boldT}{{\boldsymbol{T}}}
\newcommand{\boldTheta}{{\boldsymbol{\Theta}}}
\newcommand{\boldf}{{\boldsymbol{f}}}
\newcommand{\boldu}{{\boldsymbol{u}}}
\newcommand{\boldm}{{\boldsymbol{m}}}
\newcommand{\boldxi}{{\boldsymbol{\xi}}}
\newcommand{\boldv}{{\boldsymbol{v}}}
\newcommand{\boldk}{{\boldsymbol{k}}}
\newcommand{\boldb}{{\boldsymbol{b}}}
\newcommand{\boldbeta}{{\boldsymbol{\beta}}}
\newcommand{\boldDelta}{{\boldsymbol{\Delta}}}
\newcommand{\nnu}{\nsample}
\newcommand{\nsample}{n}
\newcommand{\subsetr}{\boldsymbol{r}}
\newcommand{\boldthetah}{{\widehat{\boldtheta}}}
\newcommand{\mathbbR}{\mathbb{R}}
\newcommand{\KL}{\mathrm{KL}}
\newcommand{\numparams}{n}
\newcommand{\boldhh}{{\widehat{\boldh}}}
\newcommand{\boldh}{{\boldsymbol{h}}}
\newcommand{\Hh}{{\widehat{H}}}
\newcommand{\boldxnu}{\boldY}
\newcommand{\boldx}{{\boldsymbol{x}}}
\newcommand{\boldxp}{{\boldsymbol{x}}^{P}}
\newcommand{\boldxq}{{\boldsymbol{x}}^{Q}}
\newcommand{\boldz}{{\boldsymbol{z}}}
\newcommand{\boldg}{{\boldsymbol{g}}}
\newcommand{\boldw}{{\boldsymbol{w}}}
\newcommand{\nde}{\nsample'}
\newcommand{\boldxde}{\boldY'}
\newcommand{\boldX}{{\boldsymbol{X}}}
\newcommand{\boldY}{{\boldsymbol{Y}}}
\newcommand{\boldy}{{\boldsymbol{y}}}
\newcommand{\boldYnu}{{\boldsymbol{Y}}}
\newcommand{\boldYde}{{\boldsymbol{Y}}}
\newcommand{\hh}{{\widehat{h}}}
\newcommand{\boldI}{{\boldsymbol{I}}}
\newcommand{\PE}{{\widehat{PE}}}
\newcommand{\ratioh}{\widehat{\ratiosymbol}}
\newcommand{\ratiosymbol}{r}
\newcommand{\ratiomodel}{g}
\newcommand{\thetah}{{\widehat{\theta}}}
\newcommand{\mathbbE}{\mathbb{E}}
\newcommand{\pnu}{p_\mathrm{te}}
\newcommand{\pde}{p_\mathrm{rf}}
\newcommand{\refsection}{\boldS_\mathrm{rf}}
\newcommand{\tesection}{\boldS_\mathrm{te}}
\newcommand{\refY}{\boldY_\mathrm{rf}}
\newcommand{\teY}{\boldY_\mathrm{te}}
\newcommand{\nseg}{n}
\newcommand{\distP}{P}
\newcommand{\distQ}{Q}
\newcommand{\iid}{\stackrel{\mathrm{i.i.d.}}{\sim}}
\newcommand{\dx}{\mathrm{d}\boldx}
\newcommand{\dy}{\mathrm{d}\boldy}

\def\ratio{r}
\def\relratio{{\ratio}_{\alpha}}

\def\ci{\perp\!\!\!\perp} 
\newcommand\independent{\protect\mathpalette{\protect\independenT}{\perp}} 
\def\independenT#1#2{\mathrel{\rlap{$#1#2$}\mkern2mu{#1#2}}} 
\newcommand*\xor{\mathbin{\oplus}}

\title{
\vspace*{-33mm}
\begin{flushleft}
  \normalsize
  \sl
   Submitted to \textit{Neural Computation}
\end{flushleft}
\vspace*{0mm}
Direct Learning of Sparse Changes in \\Markov Networks
by Density Ratio Estimation\footnote{
An earlier version of this work was presented at the
European Conference on Machine Learning and Principles and
Practice of Knowledge Discovery in Databases (ECML/PKDD2013)
on Sep. 23-27, 2013.
}
}

\author{
Song Liu \\
song@sg.cs.titech.ac.jp\\
Tokyo Institute of Technology,\\
2-12-1 O-okayama, Meguro, Tokyo 152-8552, Japan.\\
\url{http://sugiyama-www.cs.titech.ac.jp/~song/}\\[2mm]
John A. Quinn\\
jquinn@cit.ac.ug\\
Makerere University, P.O. Box 7062, Kampala, Uganda.\\[2mm]
Michael U. Gutmann\\
michael.gutmann@helsinki.fi\\
University of Helsinki, Finland, P.O. Box 68, FI-00014, Finland.\\[2mm]
Taiji Suzuki\\
suzuki.t.ct@m.titech.ac.jp\\
Tokyo Institute of Technology,\\
2-12-1 O-okayama, Meguro, Tokyo 152-8552, Japan.\\[2mm]
Masashi Sugiyama\\
sugi@cs.titech.ac.jp\\
Tokyo Institute of Technology,\\
2-12-1 O-okayama, Meguro, Tokyo 152-8552, Japan.\\
\url{http://sugiyama-www.cs.titech.ac.jp/}
}
\date{}
\maketitle
\vspace*{-5mm}
\begin{abstract}\noindent
We propose a new method for detecting changes
in Markov network structure between two sets of samples.
Instead of naively fitting two Markov network models separately
to the two data sets and figuring out their difference,
we \emph{directly} learn the network structure change
by estimating the ratio of Markov network models.
This density-ratio formulation naturally allows us to
introduce sparsity in the network structure change,
which highly contributes to enhancing interpretability.
Furthermore, computation of the normalization term,
which is a critical bottleneck of the naive approach,
can be remarkably mitigated.
We also give the dual formulation of the optimization problem, which further reduces the computation cost for large-scale Markov networks.
Through experiments,
we demonstrate the usefulness of our method.

\end{abstract} 

\section{Introduction}
\label{sec.introduction}
Changes in 
interactions between random variables are interesting in many real-world phenomena.
For example, genes may interact with each other in different ways when external stimuli change,
co-occurrence between words may appear/disappear when the domains of text corpora shift,
and
correlation among pixels may change when a surveillance camera captures anomalous activities.
Discovering such changes in interactions is a task of great interest
in machine learning and data mining,
because it provides useful insights into underlying mechanisms in many real-world applications. 

In this paper, we consider the problem of detecting changes
in conditional independence among random variables
between two sets of data.
Such conditional independence structure can be expressed
via an undirected graphical model called a \emph{Markov network} (MN)
\citep{Bishop_PRML,GM_Wainwright,PGM_Koller},
where nodes and edges represent variables and their conditional dependencies, respectively. 
As a simple and widely applicable case, the pairwise MN model has been thoroughly studied recently \citep{Ravikumar_2010,Lee_EfficientL1Learning}. Following this line, we also focus on the pairwise MN model as a representative example.

A naive approach to change detection in MNs is
the two-step procedure of first estimating two MNs separately from two sets of data
by \emph{maximum likelihood estimation} (MLE),
and then comparing the structure of the learned MNs.
However, MLE is often computationally intractable
due to the normalization factor included in the density model. 
Therefore, Gaussianity is often assumed in practice for computing
the normalization factor analytically \citep{Hastie_EMSL}, though this Gaussian assumption is 
highly restrictive in practice.
We may utilize \emph{importance sampling} \citep{RobertMCStat2005}
to numerically compute the normalization factor,
but an inappropriate choice of the instrumental distribution
may lead to an estimate with high variance 
\citep{WassermanAllStat2010};
for more discussions on sampling techniques,
see \citet{gelman1995MCSim} and \citet{hinton2002CDivergence}.
\citet{hyvarinen2005NonNormalized} and \citet{Gutmann2012} have explored 
an alternative approach to avoid computing the normalization factor
which are not based on MLE.

However, the two-step procedure has the conceptual weakness
that structure change is not directly learned.
This indirect nature causes a crucial problem:
Suppose that we want to learn a sparse structure change.
For learning sparse changes,
we may utilize $\ell_1$-regularized MLE
\citep{Banerjee_Model_Selection,Friedman_GLasso,Lee_EfficientL1Learning},
which produces sparse MNs
and thus the change between MNs also becomes sparse.
However, this approach does not work if each MN is dense
but only change is sparse.

To mitigate this indirect nature,
the \emph{fused-lasso} \citep{Tibshirani_FusedLasso}
is useful,
where two MNs are simultaneously learned
with a sparsity-inducing penalty on the \emph{difference} between two MN parameters
\citep{Gaussian_Change,Danaher2013jointGlasso}.
Although this fused-lasso approach allows us to learn sparse structure change naturally,
the restrictive Gaussian assumption is still necessary
to obtain the solution in a computationally tractable way.

The \emph{nonparanormal} assumption \citep{nonparanormal,nonparanormal_skeptic}
is a useful generalization of the Gaussian assumption.
A nonparanormal distribution is a \emph{semi-parametric Gaussian copula}
where each Gaussian variable is transformed by a monotone non-linear function.
Nonparanormal distributions are much more flexible than Gaussian distributions
thanks to the feature-wise  non-linear transformation,
while the normalization factors can still be computed analytically.
Thus, the fused-lasso method combined with nonparanormal models
would be one of the state-of-the-art approaches to change detection in MNs.
However, the fused-lasso method is still based on separate modeling of two MNs,
and its computation for more general non-Gaussian distributions is  challenging.

\begin{figure}[t]
\centering
\includegraphics[width=.35\textwidth,trim = 0mm 1.5mm 0mm 0mm, clip]{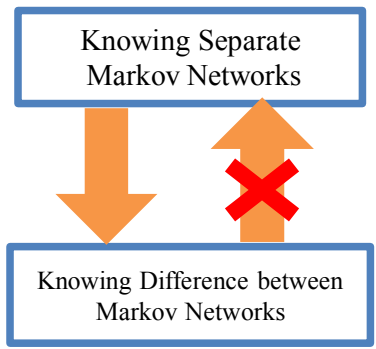}
\caption{The rationale of direct structural change learning: finding the difference between two MNs is a more specific task than finding the entire structures of those two networks, and hence should be possible to learn with less data.}
\label{fig.rationale}
\end{figure}

In this paper, we propose a more direct approach to structural change learning in MNs
based on \emph{density ratio estimation} (DRE) \citep{Density_Ratio_Book}.
Our method does not separately model two MNs,
but directly models the \emph{change} in two MNs.
This idea follows Vapnik's principle \citep{Vapnik1998}:
\begin{quote}
If you possess a restricted amount of information for solving some problem,
try to solve the problem directly and never solve a more general problem as an intermediate step.
It is possible that the available information is sufficient for a direct solution
but is insufficient for solving a more general intermediate problem.  
\end{quote}
This principle was used in the development of \emph{support vector machines} (SVMs):
rather than modeling two classes of samples,
SVM directly learns a decision boundary that is sufficient for performing
pattern recognition.
In the current context, estimating two MNs is more general than
detecting changes in MNs (Figure~\ref{fig.rationale}).
By directly detecting changes in MNs, we can also halve the number of parameters,
from two MNs to one MN-difference. 

Another important advantage of our DRE-based method
is that the normalization factor can be approximated efficiently,
because the normalization term in a density ratio function
takes the form of the expectation over a data distribution and thus it can be simply approximated by the sample average without additional sampling.
Through experiments on gene expression and Twitter data analysis,
we demonstrate the usefulness of our proposed approach.

The remainder of this paper is structured as follows.
In Section~\ref{pro.form.sec},
we formulate the problem of detecting structural changes
and review currently available approaches.
We then propose our DRE-based structural change detection method
in Section~\ref{sec.direct.learning}. 
Results of illustrative and real-world experiments are reported
in Section~\ref{sec.illustrative} and Section~\ref{sec.real.world}, respectively.
Finally, we conclude our work and show the future direction in Section~\ref{conclusion.sec}.

\section{Problem Formulation and Related Methods}
\label{pro.form.sec}
In this section, we formulate the problem of change detection
in Markov network structure and review existing approaches.

\subsection{Problem Formulation}
\label{sec.prob.form}


Consider two sets of independent samples drawn separately
from two probability distributions $P$ and $Q$ on $\mathbb{R}^d$:
\begin{align*}
 \{\boldxp_i\}_{i=1}^{n_P} \iid P \text{ and }  \{\boldxq_i\}_{i=1}^{n_Q} \iid Q.
\end{align*}
We assume that $P$ and $Q$ belong to the family of \emph{Markov networks} (MNs)
consisting of univariate and bivariate factors\footnote{
Note that the proposed algorithm itself can be applied to
\emph{any} MNs containing more than two elements in each factor.
}, i.e., 
their respective probability densities $p$ and $q$ are expressed as
\begin{align}
\label{eq.density.model}
p(\boldx;\boldtheta) =\frac{1}{Z(\boldtheta)}\exp\left(
\sum_{u,v=1, u\ge v}^{d} \boldtheta_{u,v}^\top \boldf(x^{(u)},x^{(v)}) \right),
\end{align}
where
$\boldx = (x^{(1)}, \dots, x^{(d)})^\top$ is the $d$-dimensional random variable,
$\top$ denotes the transpose,
$\boldtheta_{u,v}$ is the parameter vector for the elements $x^{(u)}$ and $x^{(v)}$, and
\begin{align*}
\boldtheta = (\boldtheta^\top_{1,1},\ldots, \boldtheta^\top_{d,1},\boldtheta^\top_{2,2},\ldots,\boldtheta^\top_{d,2},\ldots,\boldtheta^\top_{d,d})^\top
\end{align*}
is the entire parameter vector.
$\boldf(x^{(u)},x^{(v)})$ is a bivariate vector-valued basis function.
$Z(\boldtheta)$ is the normalization factor defined as
\begin{align*}
Z(\boldtheta)  = \int \exp\left(\sum_{u,v=1, u\ge v}^{d}  \boldtheta_{u,v}^\top \boldf(x^{(u)},x^{(v)})\right)\dx.
\end{align*}
$q(\boldx; \boldtheta)$ is defined in the same way.



Given two densities which can be parameterized using $p(\boldx;\boldtheta^P)$ and $q(\boldx;\boldtheta^Q)$, our goal is to discover \emph{the changes in parameters} from $P$ to $Q$, i.e., $\boldtheta^P - \boldtheta^Q$.
\subsection{Sparse Maximum Likelihood Estimation and Graphical Lasso}
\label{sec.MLE}
Maximum likelihood estimation (MLE) with group $\ell_1$-regularization
has been widely used for estimating the sparse structure of MNs
\citep{Schmidt_Convex_Log_Linear,Ravikumar_2010,Lee_EfficientL1Learning}:
\begin{align}
\label{MLE_def}
\max_\boldtheta\left[ \frac{1}{n_P}\sum_{i=1}^{n_P} \log p(\boldxp_i; \boldtheta) - \lambda \sum_{u,v=1, u\ge v}^d \|\boldtheta_{u,v}\|\right],
\end{align}
where $\|\cdot\|$ denotes the $\ell_2$-norm.
As $\lambda$ increases, $\|\boldtheta_{u,v}\|$ may drop to $0$.
Thus, this method favors an MN that encodes more conditional independencies
among variables.


Computation of the normalization term $Z(\boldtheta)$ in Eq.\eqref{eq.density.model}
is often computationally intractable
when the dimensionality of $\boldx$ is high.
To avoid this computational problem, the Gaussian assumption is often imposed
\citep{Friedman_GLasso,Meinshausen_Variable_Selection}.
More specifically, the following zero-mean Gaussian model is used:
\begin{align*}
p(\boldx;\boldTheta) = \frac{\det(\boldTheta)^{1/2}}{ {(2\pi)}^{d/2}} \exp \left( -\frac{1}{2}\boldx^\top\boldTheta\boldx \right),
\end{align*}
where $\boldTheta$ is the inverse covariance matrix (a.k.a.~the precision matrix)
and $\det(\cdot)$ denotes the determinant.
Then $\boldTheta$ is learned as
\begin{align*}
\max_\boldTheta\left[ \log \det(\boldTheta) - \mathrm{tr}(\boldTheta \boldS^P) - \lambda \|\boldTheta\|_1\right],
\end{align*}
where $\boldS^P$ is the sample covariance matrix of $\{\boldxp_i\}_{i=1}^n$.
$\|\boldTheta\|_1$ is the $\ell_1$-norm of $\boldTheta$,
i.e., the absolute sum of all elements.
This formulation has been studied intensively in \citet{Banerjee_Model_Selection}, and a computationally efficient algorithm called the \emph{graphical lasso} (Glasso)
has been proposed \citep{Friedman_GLasso}.


Sparse changes in conditional independence structure between $P$ and $Q$
can be detected by comparing two MNs estimated separately using sparse MLE.
However, this approach implicitly assumes that two MNs are sparse,
which is not necessarily true even if the change is sparse.


\subsection{Fused-Lasso (Flasso) Method}
\label{sec:fused-lasso}

To more naturally handle sparse changes in conditional independence structure between $P$ and $Q$,
a method based on \emph{fused-lasso}  \citep{Tibshirani_FusedLasso}
has been developed \citep{Gaussian_Change}.
This method directly sparsifies the \emph{difference} between parameters. 


The original method conducts \emph{feature-wise neighborhood regression} \citep{Meinshausen_Variable_Selection} jointly for $P$ and $Q$,
which can be conceptually understood as maximizing the local conditional Gaussian likelihood jointly on each feature \citep{Ravikumar_2010}.
A slightly more general form of the learning criterion may be summarized as
\begin{align*}
\max_{\boldtheta_{s}^P, \boldtheta_{s}^Q} \left[ \ell_s^P(\boldtheta^P_s) + \ell_s^Q(\boldtheta^Q_s)
- \lambda_1 (\|\boldtheta^P_s\|_1+\|\boldtheta^Q_s\|_1) - \lambda_2
\|\boldtheta_{s}^P-\boldtheta_{s}^Q\|_1\right],
\end{align*}
where
$\ell_s^P(\boldtheta)$ is the log conditional likelihood
for the $s$-th element $x^{(s)}\in\mathbbR$ given the rest $\boldx^{(-s)} \in\mathbbR^{d-1}$:
\begin{align*}
\ell_s^P(\boldtheta) = \frac{1}{n_P}\sum_{i=1}^{n_P}\log p(x_{i}^{(s)}{}^{P}|\boldx^{(-s)}_{i}{}^{P};\boldtheta).
\end{align*} 
$\ell_s^Q(\boldtheta)$ is defined in the same way as $\ell_s^P(\boldtheta)$. 

Since the Flasso-based method directly sparsifies the change in MN structure,
it can work well even when each MN is not sparse.
However, using other models than Gaussian is difficult
because of the normalization issue described in Section~\ref{sec.MLE}.



\subsection{Nonparanormal Extensions}
\label{sec:nonparanormal}
In the above methods, Gaussianity is required in practice
to compute the normalization factor efficiently,
which is a highly restrictive assumption.
To overcome this restriction,
it has become popular to perform structure learning under the \emph{nonparanormal} settings
\citep{nonparanormal,nonparanormal_skeptic},
where the Gaussian distribution is replaced by a \emph{semi-parametric Gaussian copula}.

A random vector $\boldx = (x^{(1)}, \dots, x^{(d)})^\top$
is said to follow a \emph{nonparanormal} distribution,
if there exists a set of monotone and differentiable functions, $\{h_i(x)\}_{i=1}^d$, such that 
$\boldh(\boldx) = (h_1(x^{(1)}), \ldots, h_d(x^{(d)}))^\top$
follows the Gaussian distribution.
Nonparanormal distributions are much more flexible than Gaussian distributions
thanks to the non-linear transformation $\{h_i(x)\}_{i=1}^d$,
while the normalization factors can still be computed in an analytical way.

However, the nonparanormal transformation is restricted to be element-wise,
which is still restrictive to express complex distributions.

\subsection{Maximum Likelihood Estimation for Non-Gaussian Models by Importance-Sampling}
\label{sec.ismle}
A numerical way to obtain the MLE solution under general non-Gaussian distributions 
is \emph{importance sampling}.

Suppose that we try to maximize the log-likelihood\footnote{From here on, we simplify $\sum_{u,v=1, u\ge v}^{d}$ as $\sum_{u\ge v}$.}:
\begin{align}
\label{eq.ismle.obj}
\ell_\text{MLE}(\boldtheta) &= \frac{1}{n_P} \sum_{i=1}^{n_P} \log p(\boldx^P_i;\boldtheta)\notag\\ 
&=\frac{1}{n_P} \sum_{i=1}^{n_P} \sum_{u\ge v} \boldtheta_{u,v}^\top \boldf(x^{(u)P}_i,x^{(v)P}_i) - \log\int \exp\left(\sum_{u\ge v} \boldtheta_{u,v}^\top \boldf(x^{(u)},x^{(v)})\right)\;\dx.
\end{align}

The key idea of importance sampling is
to compute the integral
by the expectation over an easy-to-sample \emph{instrumental density} $p'(\boldx)$
(e.g., Gaussian)
weighted according to the \emph{importance} $1/p'(\boldx)$.
More specifically, 
using i.i.d.~samples $\{\boldx'_i\}_{i=1}^{n'} \iid p'(\boldx)$,
the last term of Eq.\eqref{eq.ismle.obj} can be approximately computed 
as follows:
\begin{align*}
\log\int \exp\left(\sum_{u\ge v} \boldtheta_{u,v}^\top \boldf(x^{(u)},x^{(v)})\right)\;\dx 
&=\log\int p'(\boldx) \frac{\exp\left(\sum_{u\ge v} \boldtheta_{u,v}^\top \boldf(x^{(u)},x^{(v)})\right)}{p'(\boldx)}\;\dx\\
&\approx
\log\frac{1}{n'} \sum_{i=1}^{n'} \frac{\exp\left(\sum_{u\ge v} \boldtheta_{u,v}^\top \boldf(x'^{(u)}_i,x'^{(v)}_i)\right) }{p'(\boldx'_i)}.
\end{align*}
We refer to this implementation of Glasso as IS-Glasso below.

However, importance sampling tends to produce an estimate with large variance
if the instrumental distribution is not carefully chosen.
Although it is often suggested to use a density whose shape is similar to the function to be integrated but with thicker tails as $p'$,
it is not straightforward in practice to decide which $p'$ to choose,
especially when the dimensionality of $\boldx$ is high \citep{WassermanAllStat2010}.
 
We can also consider an importance-sampling version of the Flasso method
(which we refer to as IS-Flasso)\footnote{For implementation simplicity, we maximize the joint likelihood of $p$ and $q$, instead of its feature-wise conditional likelihood. We also switch the first penalty term from $\ell_1$ to $\ell_2$. 
} 
\begin{align*}
\max_{\boldtheta^P, \boldtheta^Q} \left[ \ell_{\text{MLE}}^P(\boldtheta^P) + \ell_{\text{MLE}}^Q(\boldtheta^Q)
- \lambda_1 (\|\boldtheta^P\|^2+\|\boldtheta^Q\|^2) - \lambda_2
\sum_{u\ge v}\|\boldtheta_{u,v}^P-\boldtheta_{u,v}^Q\|\right],
\end{align*}
where both $\ell_{\text{MLE}}^P(\boldtheta^P) $ and $ \ell_{\text{MLE}}^Q(\boldtheta^Q)$ are approximated by importance sampling for non-Gaussian distributions.
However, in the same way as IS-Glasso,
the choice of instrumental distributions is not straightforward.




\section{Direct Learning of Structural Changes via Density Ratio Estimation}
\label{sec.direct.learning}
The Flasso method can more naturally handle sparse changes in MNs
than separate sparse MLE.
However, the Flasso method is still based on separate modeling of two MNs,
and its computation for general high-dimensional non-Gaussian distributions is challenging.
In this section, we propose to directly learn structural changes
based on \emph{density ratio estimation} \citep{Density_Ratio_Book}.
Our approach does not involve separate modeling of each MN
and allows us to approximate the normalization term efficiently
for \emph{any} distributions.

\subsection{Density Ratio Formulation for Structural Change Detection}
Our key idea is to consider the ratio of $p$ and $q$:

\begin{align*}
\frac{p(\boldx; \boldthetaP)}{q(\boldx; \boldthetaQ)}
\propto \exp \left( \sum_{u\ge v} (\boldthetaP_{u,v}-\boldthetaQ_{u,v})^\top \boldf(x^{(u)},x^{(v)})\right).
\end{align*}
Here $\boldthetaP_{u,v} - \boldthetaQ_{u,v}$ encodes the difference
between $\distP$ and $\distQ$ for factor $\boldf(x^{(u)},x^{(v)})$, i.e., 
$\boldthetaP_{u,v} - \boldthetaQ_{u,v}$ is zero
if there is no change in the factor $\boldf(x^{(u)},x^{(v)})$.




Once we consider the ratio of $p$ and $q$, we actually do not have to estimate
$\boldthetaP_{u,v}$ and $\boldthetaQ_{u,v}$;
instead estimating their difference $\boldtheta_{u,v}=\boldthetaP_{u,v} - \boldthetaQ_{u,v}$
is sufficient for change detection:
\begin{align}
\label{ratio.model.def}
r(\boldx;\boldtheta) =
\frac{1}{N(\boldtheta)} \exp\left(\sum_{u\ge v} \boldtheta_{u,v}^\top \boldf(x^{(u)},x^{(v)})\right),
\end{align}
where
\begin{align*}
N(\boldtheta) = \int q(\boldx) \exp\left(\sum_{u\ge v} \boldtheta_{u,v}^\top \boldf(x^{(u)},x^{(v)}) \right)\dx.  
\end{align*}
The normalization term $N(\boldtheta)$ guarantees\footnote{
If the model $q(\boldx;\boldtheta^Q)$ is correctly specified,
i.e., there exists ${\boldtheta^Q}^*$ such that $q(\boldx;{\boldtheta^Q}^*) = q(\boldx)$, 
then $N(\boldtheta)$ can be interpreted
as importance sampling of
$Z(\boldtheta^P)$ via instrumental distribution $q(\boldx)$. Indeed, since
\begin{align*}
Z(\boldtheta^P) = \int q(\boldx) \frac{\exp\left(\sum_{u\ge v} {\boldtheta_{u,v}^P}^\top\boldf(x^{(u)},x^{(v)})\right)}{q(\boldx;{\boldtheta^Q}^*)} \dx,
\end{align*}
where $q(\boldx;{\boldtheta^Q}^*) = q(\boldx)$, we have
\begin{align*}
N(\boldtheta^P - {\boldtheta^Q}^*) = \frac{Z(\boldtheta^P)}{Z({\boldtheta^Q}^*)} = \int q(\boldx) \exp\left(\sum_{u\ge v} {(\boldtheta_{u,v}^P-{\boldtheta_{u,v}^Q}^*)}^\top\boldf(x^{(u)},x^{(v)})\right) \dx.
\end{align*}
This is exactly the normalization term $N(\boldtheta)$
of the ratio $ {p(\boldx;\boldtheta^P)}/{q(\boldx;{\boldtheta^Q}^*)}$. 
However, we note that the density ratio estimation method we use in this paper is consistent
to the optimal solution in the model
even without the correct model assumption \citep{Density_Ratio_Analysis}.
An alternative normalization term,
\[
N'(\boldtheta,\boldtheta^Q) = \int q(\boldx;\boldtheta^Q) r(\boldx;\boldtheta)\dx,
\]
may also be considered, as in the case of MLE.
However,
this alternative form requires an extra parameter $\boldtheta^Q$
which is not our main interest.
}
\begin{align*}
  \int q(\boldx)r(\boldx;\boldtheta) \dx = 1.
\end{align*}

Thus, in this density ratio formulation,
we are no longer modeling $p$ and $q$ separately,
but we model the change from $p$ to $q$ \emph{directly}.
This direct nature would be more suitable for change detection purposes
according to Vapnik's principle
that encourages avoidance of solving more general problems as an intermediate step
\citep{Vapnik1998}.
This direct formulation also allows us to halve the number of parameters
from both $\boldthetaP$ and $\boldthetaQ$ to only $\boldtheta$.

Furthermore, the normalization factor $N(\boldtheta)$ in the density ratio formulation can be easily 
approximated by the sample average over $\{\boldxq_i\}_{i=1}^{n_Q}\iid q(\boldx)$,
because $N(\boldtheta)$ is the expectation over $q(\boldx)$:
\begin{align*}
  N(\boldtheta) \approx
\frac{1}{n_Q}\sum_{i=1}^{n_Q}
 \exp\left(\sum_{u\ge v} \boldtheta_{u,v}^\top \boldf(x^{(u)Q}_i, x^{(v)Q}_i) \right).
\end{align*}



\subsection{Direct Density-Ratio Estimation} 
\label{estimate.ratio.sec}
Density ratio estimation has been recently introduced
to the machine learning community 
and is proven to be useful in a wide range of applications
\citep{Density_Ratio_Book}.
Here, we concentrate on the density ratio estimator called 
the \emph{Kullback-Leibler importance estimation procedure} (KLIEP) 
for log-linear models \citep{Covariate_Shift,Log_Linear_KLIEP}.


For a density ratio model $r(\boldx; \boldtheta)$,
the KLIEP method minimizes the Kullback-Leibler divergence
from $p(\boldx)$ to $\widehat{p}(\boldx) = q(\boldx) r(\boldx;\boldtheta)$:
\begin{align}
\KL[p\|\widehat{p}] 
&= \int p(\boldx) \log\frac{p(\boldx)}{q(\boldx)r(\boldx;\boldtheta)} \dx\nonumber\\
&=\text{Const.} - \int p(\boldx) \log r(\boldx; \boldtheta) \dx.
\label{eq.obj}
\end{align}
Note that our density-ratio model \eqref{ratio.model.def} automatically
satisfies the non-negativity and normalization constraints:
\begin{align*}
r(\boldx;\boldtheta) \ge 0
~~\mbox{and}~~
 \int q(\boldx) r(\boldx; \boldtheta) \dx = 1.
\end{align*}
In practice, we maximize
the empirical approximation of the second term in Eq.\eqref{eq.obj}:
\begin{align*}
\ell_{\mathrm{KLIEP}}(\boldtheta) &= 
\frac{1}{n_P}\sum_{i=1}^{n_P} \log r(\boldxp_i; \boldtheta)\\
&=  \frac{1}{n_P}\sum_{i=1}^{n_P} \sum_{u\ge v} \boldtheta_{u,v}^\top \boldf(x^{(u)P}_i,x^{(v)P}_i)
 \\
&~~~-\log \left(\frac{1}{n_Q}\sum_{i=1}^{n_Q} \exp\left(\sum_{u\ge v} \boldtheta_{u,v}^\top \boldf(x^{(u)Q}_i,x^{(v)Q}_i)\right)\right).
\end{align*}
Because $\ell_{\mathrm{KLIEP}}(\boldtheta)$ is concave with respect to $\boldtheta$,
its global maximizer can be numerically found by standard optimization techniques
such as gradient ascent or quasi-Newton methods.
The gradient of $\ell_{\mathrm{KLIEP}}$ with respect to $\boldtheta_{u,v}$ is given by
\begin{align*}
\nabla_{\boldtheta_{u,v}} \ell_{\text{KLIEP}}(\boldtheta)
=&
\frac{1}{n_P}\sum_{i=1}^{n_P} \boldf (\boldx^{(u)P}_i,\boldx^{(v)P}_i)\\
&-  \frac{\frac{1}{n_Q}\sum_{i=1}^{n_Q}
\exp \left({\sum_{u'\ge v'} \boldtheta_{u',v'}^\top \boldf(x^{(u')Q}_i,x^{(v')Q}_i)}\right) \boldf(x^{(u)Q}_i,x^{(v)Q}_i)}
{\frac{1}{n_Q}\sum_{j=1}^{n_Q}\exp\left(\sum_{u''\ge v''} \boldtheta_{u'',v''}^\top \boldf(x^{(u'')Q}_{j},x^{(v'')Q}_{j}) \right)},
\end{align*}
which can be computed in a straightforward manner for \emph{any} feature vector $\boldf(x^{(u)},x^{(v)})$.


\subsection{Sparsity-Inducing Norm}
\label{sec.sparse}
To find a sparse change between $P$ and $Q$,
we propose to regularize the KLIEP solution
with a sparsity-inducing norm $\sum_{u\ge v} \| \boldtheta_{u,v} \|$.
Note that 
the MLE approach sparsifies both $\boldtheta^P$ and $\boldtheta^Q$
so that the difference $\boldtheta^P-\boldtheta^Q$ is also sparsified,
while we directly sparsify the difference $\boldtheta^P-\boldtheta^Q$;
thus our method can still work well even if $\boldtheta^P$ and $\boldtheta^Q$ are dense.


In practice, we may use the following \emph{elastic-net} penalty \citep{JRSS-B:Zou+Hastie:2005}
to better control overfitting to noisy data:
\begin{align}
\label{eq.final.obj}
\max_\boldtheta \left[ \ell_{\text{KLIEP}}(\boldtheta) - \lambda_1 \|\boldtheta\|^2 - \lambda_2 \sum_{u\ge v} \| \boldtheta_{u,v} \| \right],
\end{align}
where $\|\boldtheta\|^2$ penalizes the magnitude of the entire parameter vector. 

\subsection{Dual Formulation for High-Dimensional Data}
\begin{figure}[t]
\centering
\includegraphics[width = .8\textwidth]{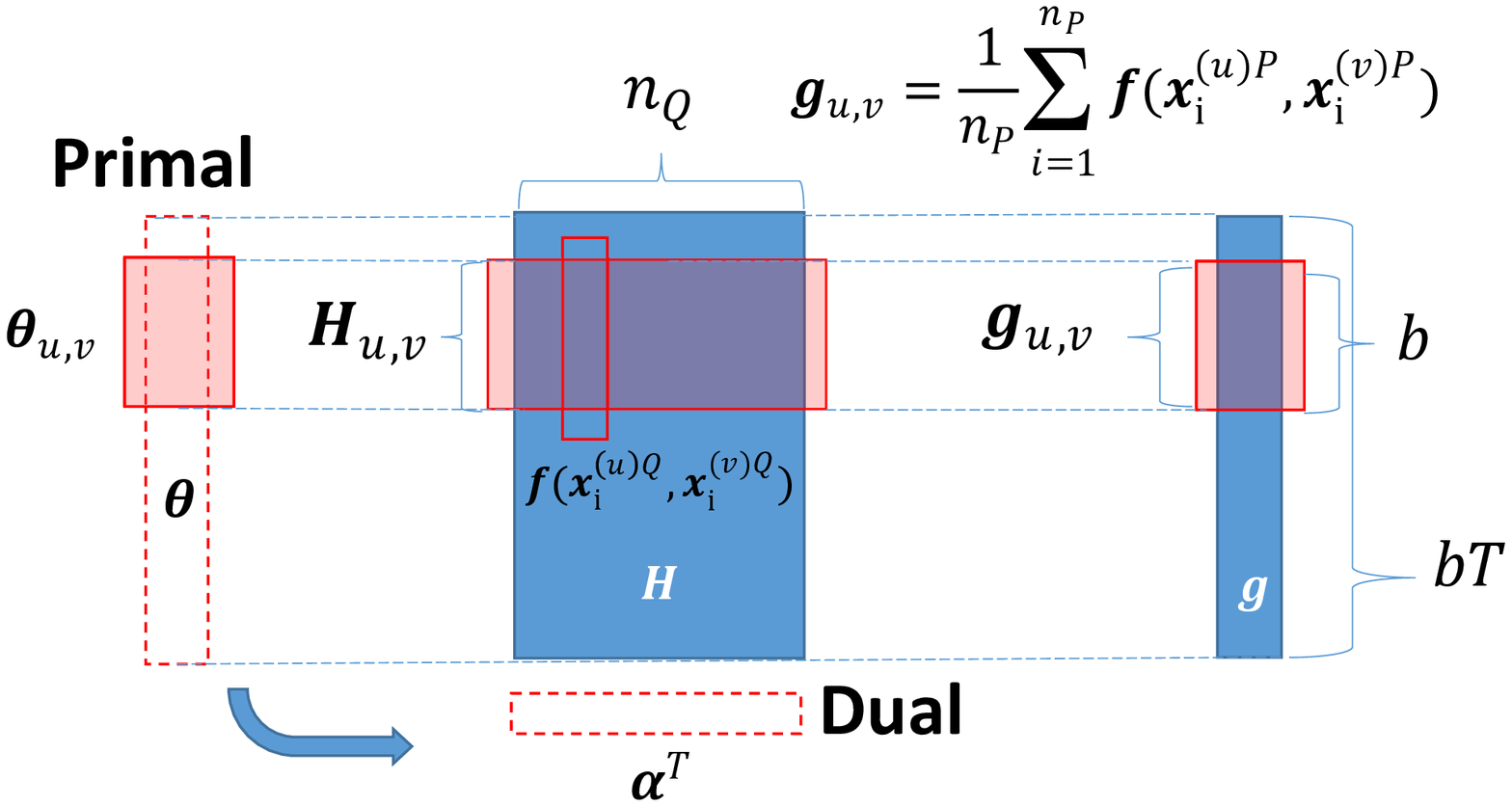}
\caption{Schematics of primal and dual optimization.
$b$ denotes the number of basis functions
and $T$ denotes the number of factors.
Because we are considering pairwise factors,
$T={\mathcal O}(d^2)$ for input dimensionality $d$.
}
\label{fig.dual.illus}
\end{figure}

The solution of the optimization problem \eqref{eq.final.obj} can
be easily obtained by standard sparse optimization methods. However, in the case where the input dimensionality $d$ is high
(which is often the case in our setup), the dimensionality of
parameter vector $\boldtheta$ is large, and thus obtaining the
solution can be computationally expensive. 
Here, we derive a dual optimization problem \citep{book:Boyd+Vandenberghe:2004},
which can be solved more efficiently for high-dimensional $\boldtheta$
(Figure~\ref{fig.dual.illus}).

As detailed in Appendix,
the dual optimization problem is given as
\begin{align}
&\min_{\boldalpha=(\alpha_1,\ldots,\alpha_{n_Q})^\top}
 \sum_{i=1}^{n_Q} \alpha_i\log \alpha_i 
 + \frac{1}{\lambda_1}\sum_{u\ge v} \max(0, \|\boldxi_{u,v}\| - \lambda_2)^2
\nonumber\\
&\text{subject to } \alpha_1,\ldots,\alpha_{n_Q}\ge0
\text{ and }\sum_{i=1}^{n_Q} \alpha_i = 1,
\label{eq.obj.dual}
\end{align}
where
\begin{align*}
\boldxi_{u,v} &=\boldg_{u,v}-\boldH_{u,v}\boldalpha,\\
\boldH_{u,v} &= [\boldf(x_1^{(u)Q},x_1^{(v)Q}), \ldots, \boldf(x_{n_Q}^{(u)Q},x_{n_Q}^{(v)Q})],\\
\boldg_{u,v} &= \frac{1}{n_P}\sum_{i=1}^{n_P} \boldf(x_i^{(u)P},x_i^{(v)P}).
\end{align*}
The primal solution can be obtained from the dual solution as
\begin{align}
\label{eq.dual.solution}
\boldtheta_{u,v} &= 
\begin{cases}
\displaystyle
\frac{1}{\lambda_1}\left( 1- \frac{\lambda_2}{\|\boldxi_{u,v}\|}\right)\boldxi_{u,v}& \text{if } \|\boldxi_{u,v}\|> \lambda_2, \\[4mm]
\displaystyle
\boldsymbol{0} &\text{if } \|\boldxi_{u,v}\| \le \lambda_2.
\end{cases}
\end{align}


Note that the dimensionality of the dual variable $\boldalpha$ is equal to $n_Q$,
while that of $\boldtheta $ is quadratic with respect to the input dimensionality $d$,
because we are considering pairwise factors.
Thus, if $d$ is not small and $n_Q$ is not very large
(which is often the case in our experiments shown later),
solving the dual optimization problem would be computationally more efficient.
Furthermore, the dual objective (and its gradient) can be computed efficiently
in parallel for each $(u,v)$,
which is a useful property when handling large-scale MNs.
Note that the dual objective is differentiable everywhere,
while the primal objective is not.

\section{Numerical Experiments}
\label{sec.illustrative}
In this section, we compare the performance of the proposed KLIEP-based method,
the Flasso method, and the Glasso method
for Gaussian models, nonparanormal models, and non-Gaussian models.
Results are reported on datasets with three different underlying distributions:
multivariate Gaussian, nonparanormal, and non-Gaussian ``diamond'' distributions.
We also investigate the computation time of the primal and dual formulations
as a function of the input dimensionality. The MATLAB implementation of the primal and dual methods are available at
\begin{center}
\url{http://sugiyama-www.cs.titech.ac.jp/~song/SCD.html}.
\end{center}




\subsection{Gaussian Distribution}
\label{sec.toy.gaussian}

First, we investigate the performance of each method under Gaussianity.


Consider a $40$-node sparse Gaussian MN, where its graphical structure is characterized by precision matrix $\boldTheta^{P}$ with diagonal elements equal to $2$. The off-diagonal elements are randomly chosen\footnote{We set $\Theta_{u,v}=\Theta_{v,u}$ for not breaking the symmetry of the precision matrix.} and set to $0.2$, so that the overall sparsity of $\boldTheta^P$ is $25\%$. We then introduce changes by randomly picking $15$ edges and reducing the corresponding elements in the precision matrix by $0.1$. The resulting precision matrices $\boldTheta^P$ and $\boldTheta^Q$ are used for drawing samples as
\begin{align*}
\{\boldx_i^P\}_{i=1}^{n_P}\iid \mathcal{N}(\boldsymbol{0},(\boldTheta^P)^{-1})
 ~~\mbox{and}~~
\{\boldx_i^Q\}_{i=1}^{n_Q} \iid \mathcal{N}(\boldsymbol{0},(\boldTheta^Q)^{-1}),
\end{align*} 
where $\mathcal{N}(\boldsymbol{\mu},\boldsymbol{\Sigma})$
denotes the multivariate normal distribution with mean $\boldsymbol{\mu}$
and covariance matrix $\boldsymbol{\Sigma}$.
Datasets of size $n = n_P=n_Q=50, 100$ are tested.

\begin{figure*}[t]
\centering
\subfigure[KLIEP, $n = 100$]{
\includegraphics[width = .31\textwidth]{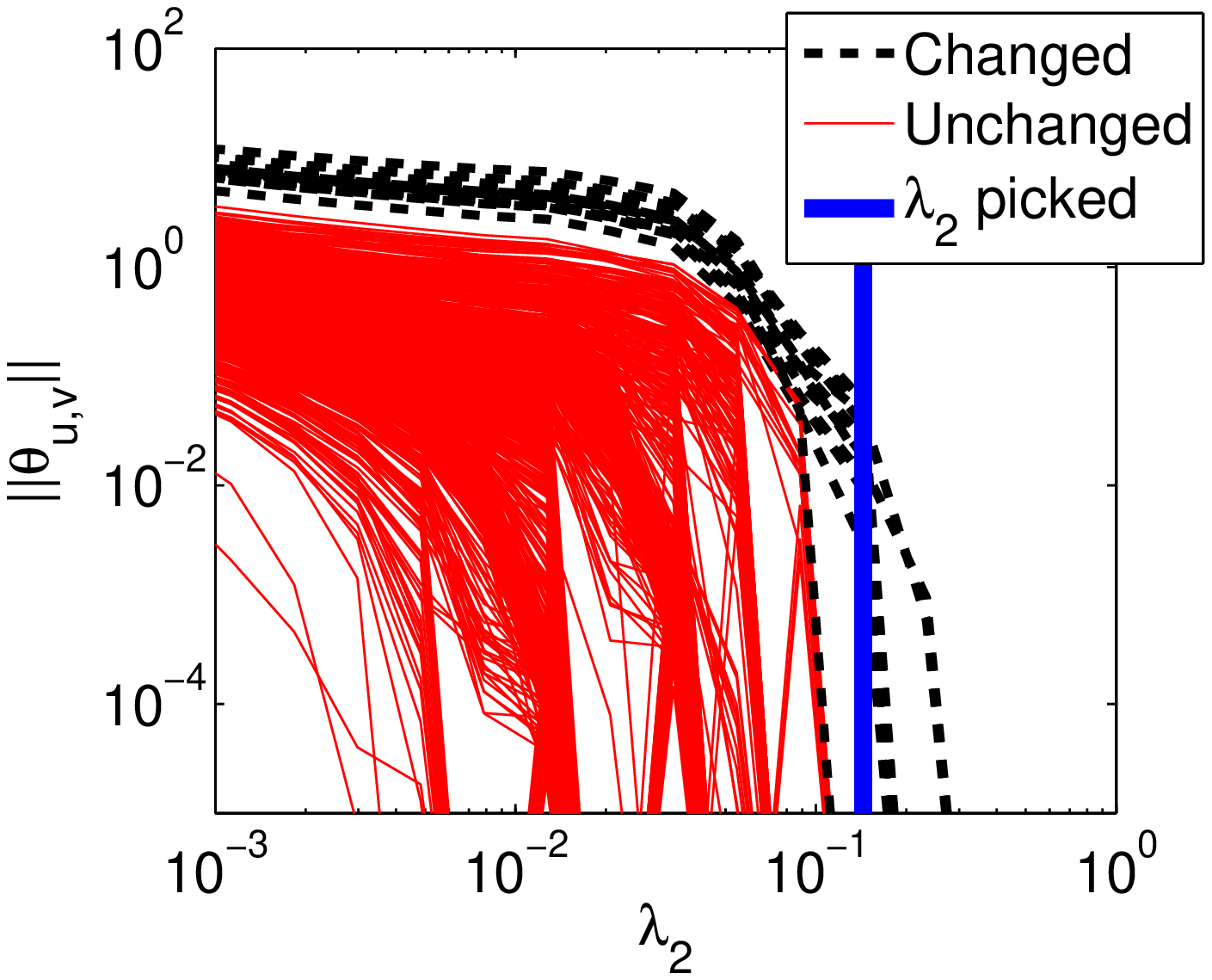}
\label{fig.toy.gaussian100.KLIEP}
}
\subfigure[Flasso, $n = 100$]{
\includegraphics[width = .31\textwidth]{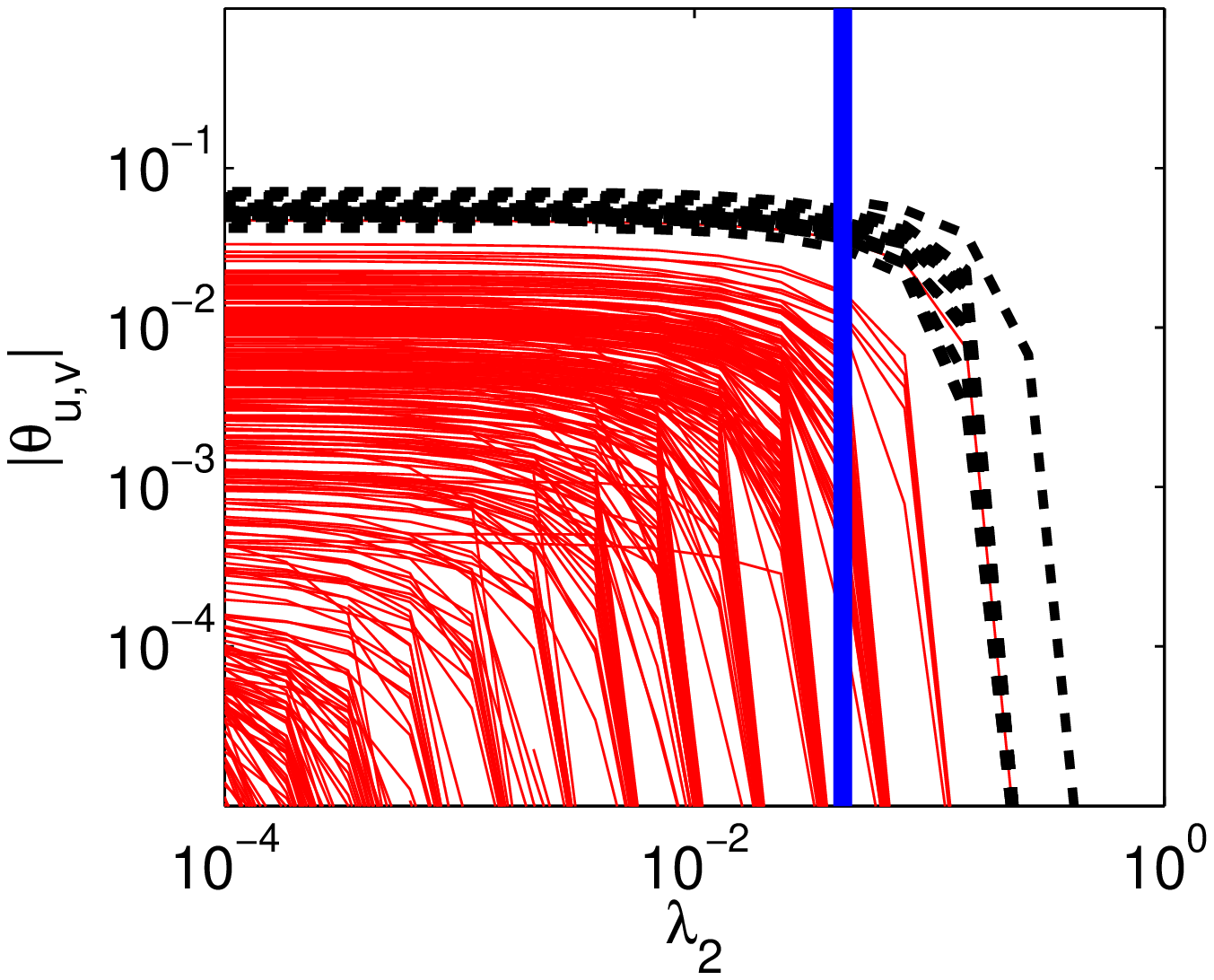}
\label{fig.toy.gaussian100.FL}
}
\subfigure[Glasso, $n = 100$]{
\includegraphics[width = .31\textwidth]{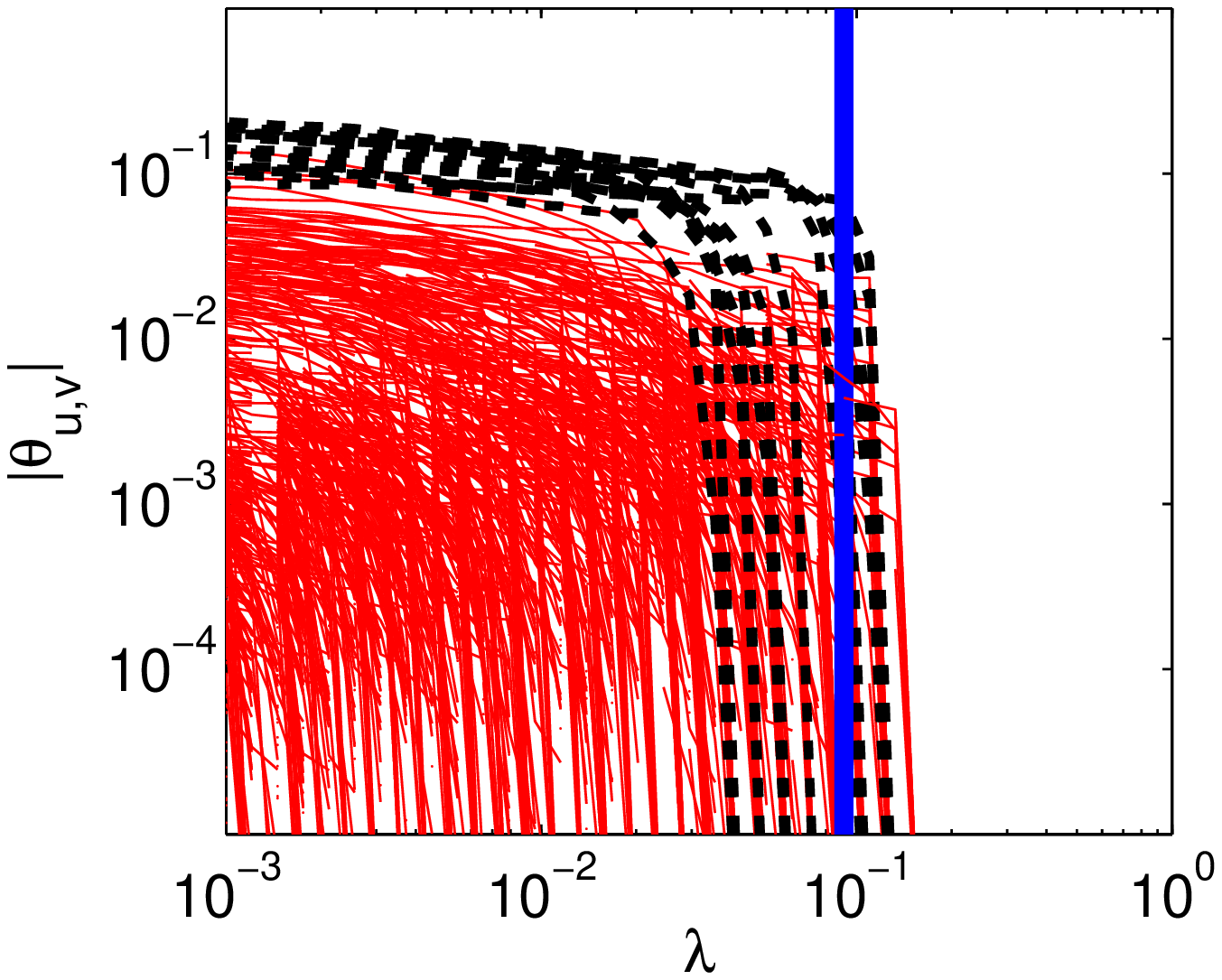}
\label{fig.toy.gaussian100.GL}
}
\subfigure[KLIEP, $n = 50$]{
\includegraphics[width = .31\textwidth]{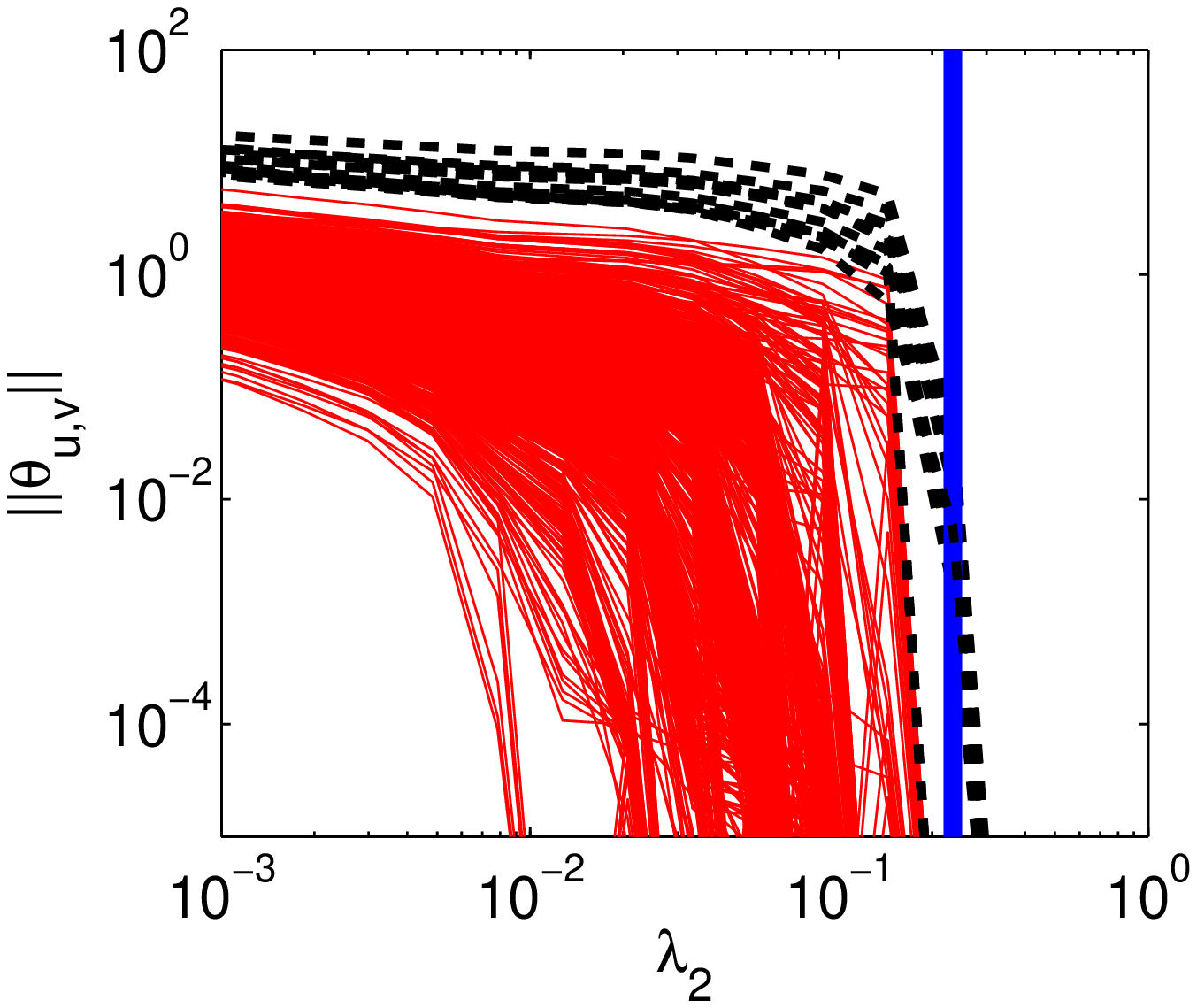}
\label{fig.toy.gaussian50.KLIEP}
}
\subfigure[Flasso, $n = 50$]{
\includegraphics[width = .31\textwidth]{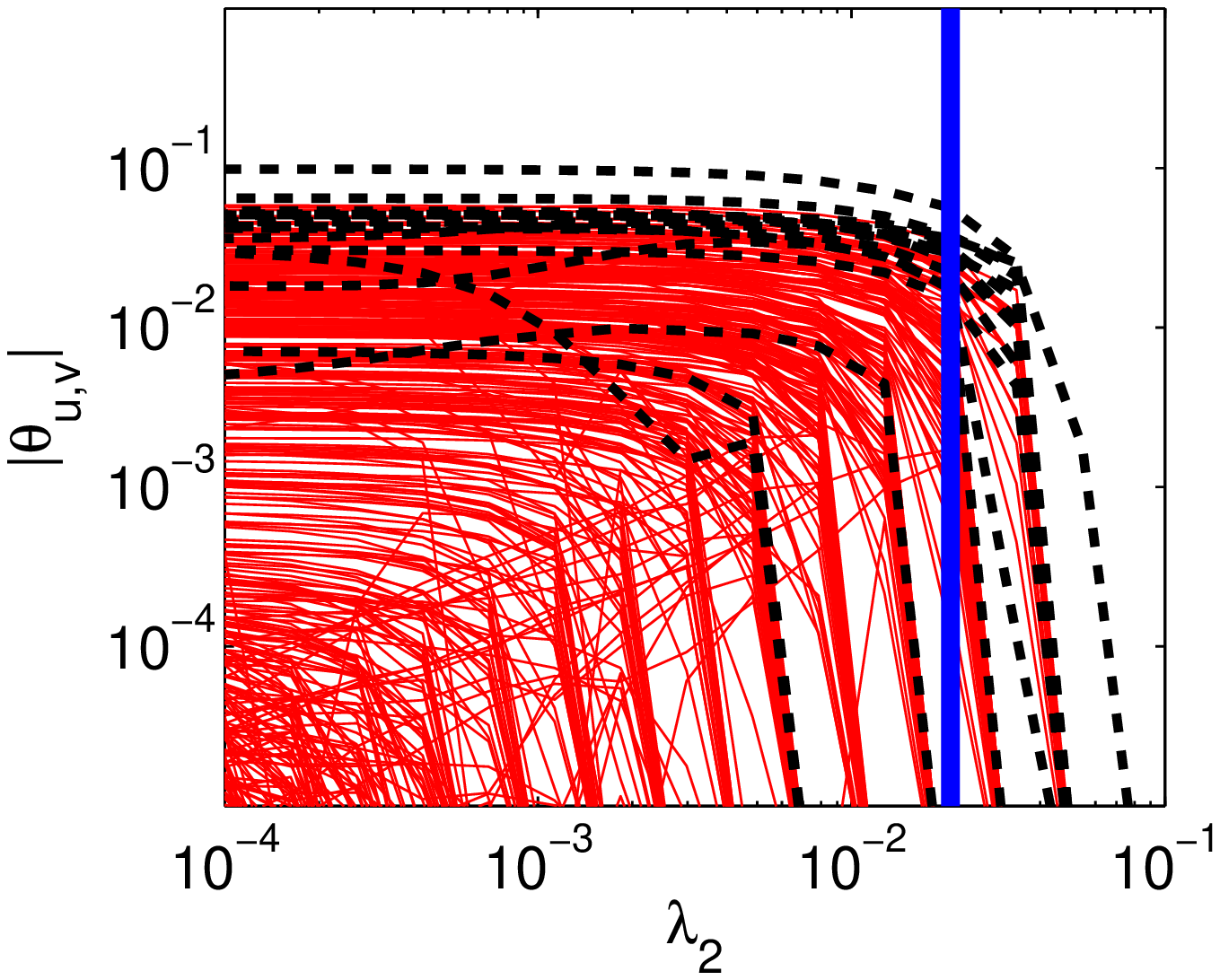}
\label{fig.toy.gaussian50.FL}
}
\subfigure[Glasso, $n =50$]{
\includegraphics[width = .31\textwidth]{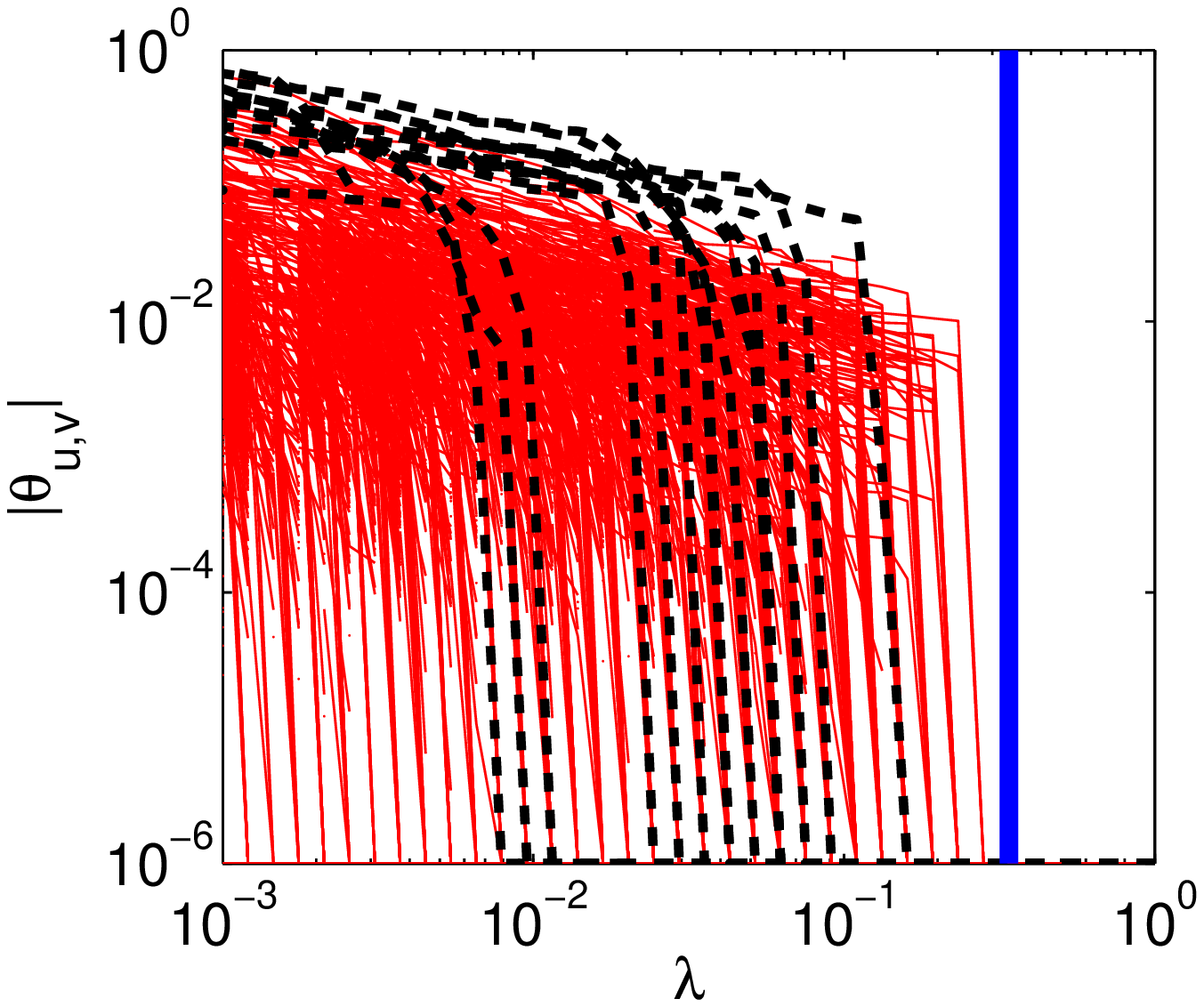}
\label{fig.toy.gaussian50.GL}
}
\subfigure[Gaussian distribution]{
\includegraphics[width = .31\textwidth]{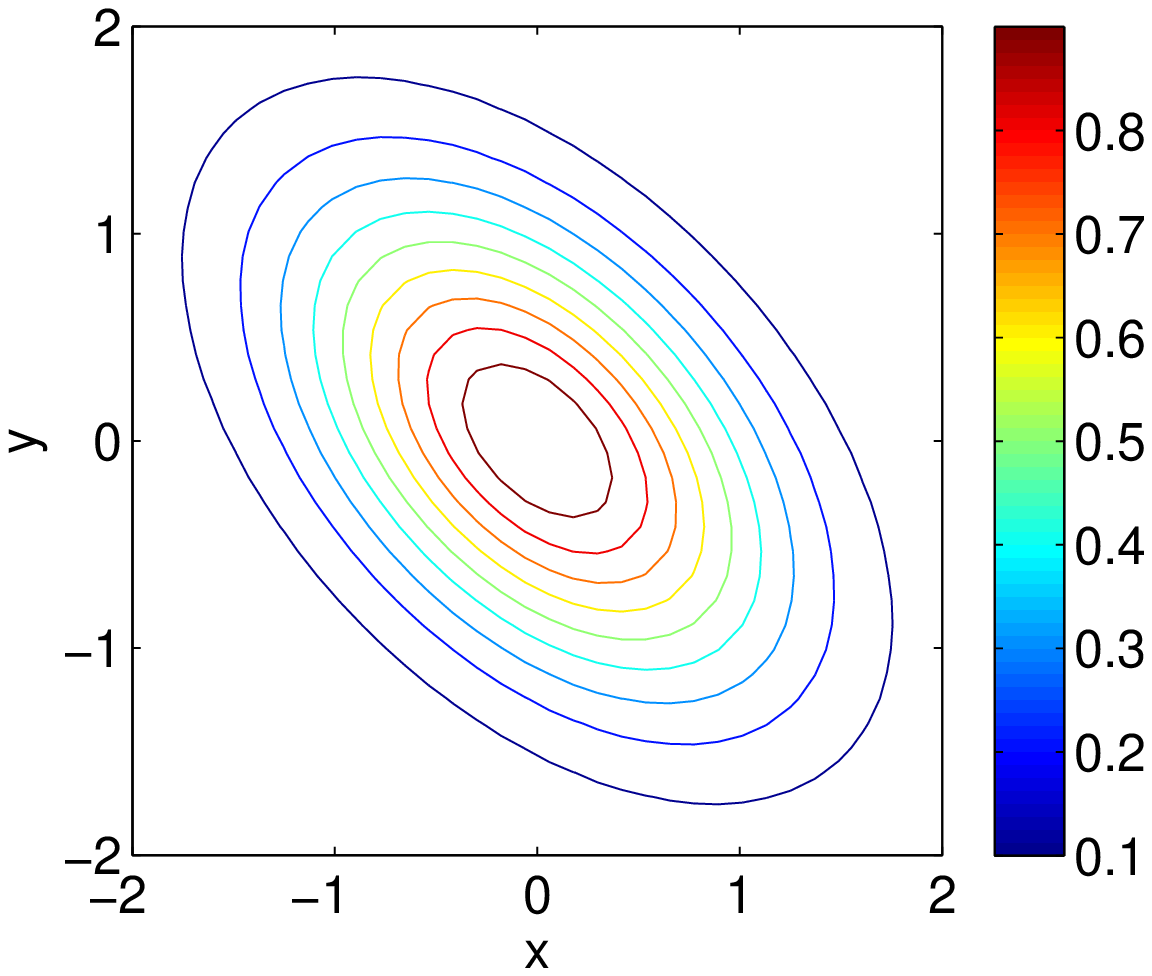}
\label{fig.toy.gaussian.contour}
}
\subfigure[P-R curve, $n =100$ ]{
\includegraphics[width = .31\textwidth]{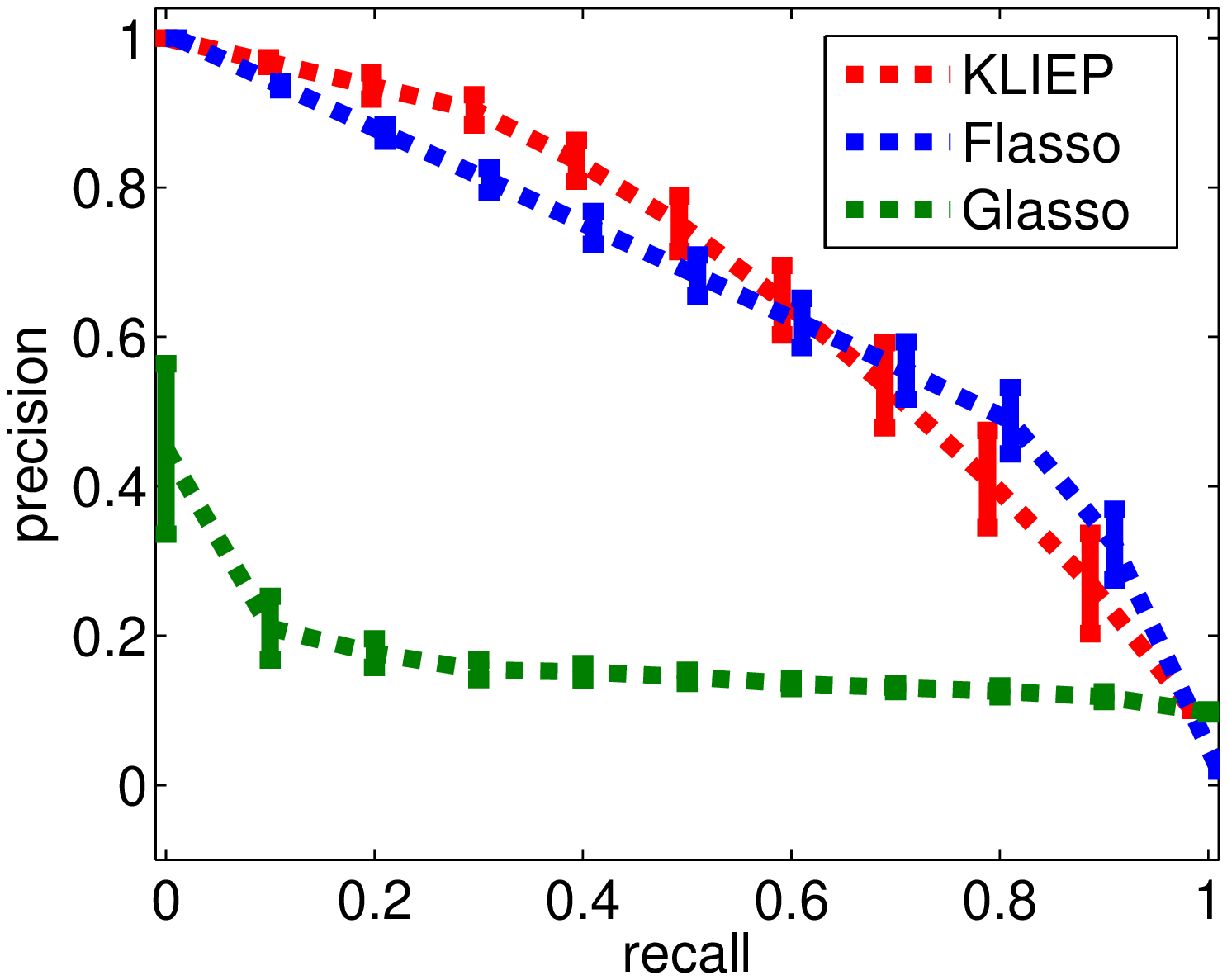}
\label{fig.toy.gaussian100.PR}
}
\subfigure[P-R curve, $n =50$ ]{
\includegraphics[width = .31\textwidth]{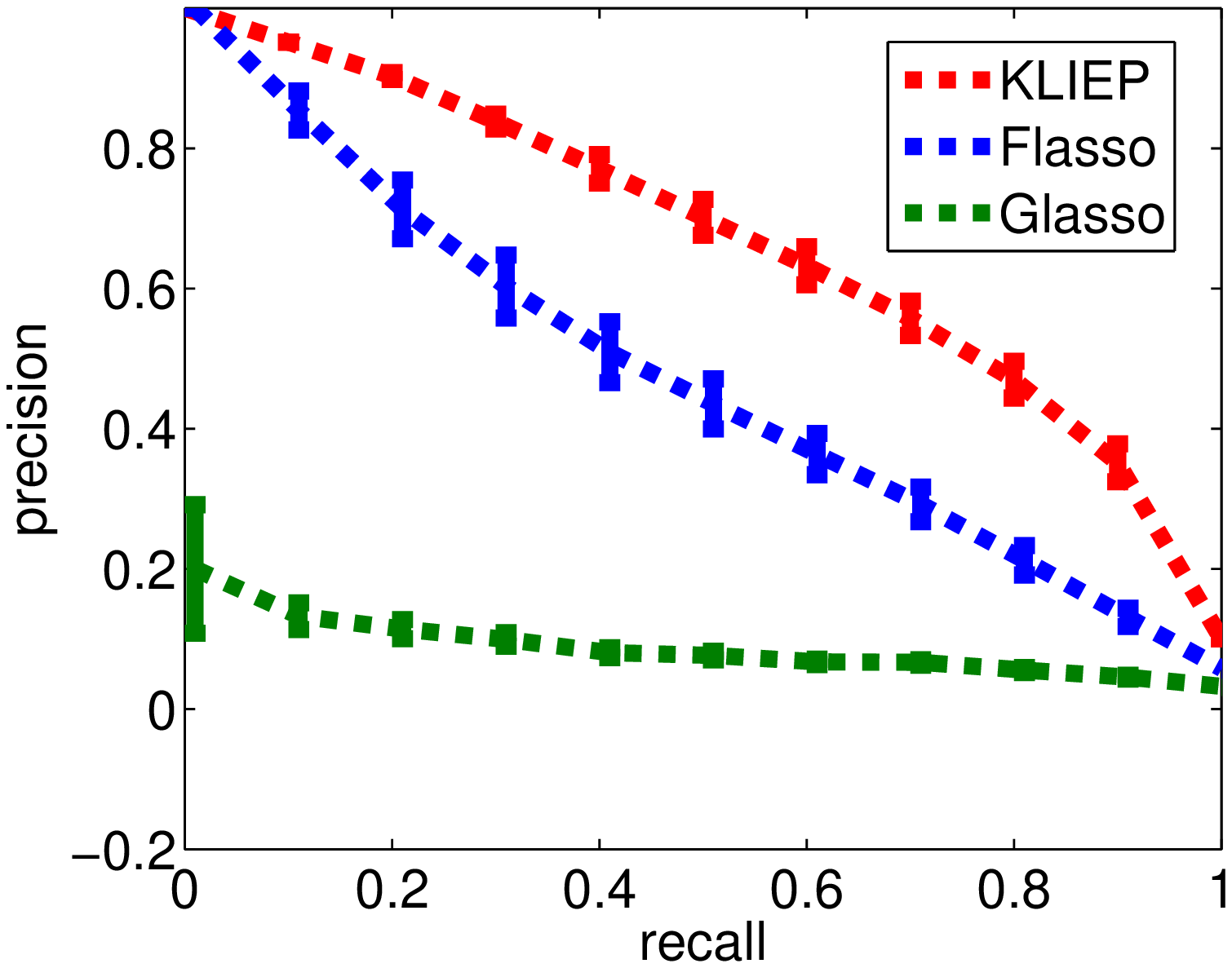}
\label{fig.toy.gaussian50.PR}
}
\caption{Experimental results on the Gaussian dataset.}
\label{fig.toy.gaussian}
\end{figure*}

We compare the performance of the KLIEP, Flasso, and Glasso methods.
Because all methods use the same Gaussian model,
the difference in performance is caused only by
the difference in estimation methods.
We repeat the experiments $20$ times with randomly generated datasets
and report the results in Figure~\ref{fig.toy.gaussian}.

The top $6$ graphs are examples of regularization paths\footnote{Paths of univariate factors are omitted for clear visibility.}.
The dashed lines represent changed edges in the ground truth,
while the solid lines represent unchanged edges. 
The top row is for $n = 100$ while the middle row is for  $n  = 50$.
The bottom $3$ graphs are the data generating distribution and averaged
precision-recall (P-R) curves with standard error over $20$ runs.
The P-R curves are plotted 
by varying the group-sparsity control parameter $\lambda_2$ with $\lambda_1=0$
in KLIEP and Flasso,
and by varying the sparsity control parameters as $\lambda=\lambda^P=\lambda^Q$
in Glasso.

In the regularization path plots,
solid vertical lines show
the regularization parameter values picked
based on hold-out data
$\{\widetilde{\boldx}_i^P\}_{i=1}^{3000} \iid P$ and
$\{\widetilde{\boldx}_i^Q\}_{i=1}^{3000} \iid Q$
as follows:
\begin{itemize}
\item \textbf{KLIEP:}
The \emph{hold-out log-likelihood} (HOLL) is maximized:
\begin{align*}
 \frac{1}{\widetilde{n}_P}\sum_{i=1}^{\widetilde{n}_P}
\log
 \frac{\exp\left(\sum_{u\ge v} \widehat{\boldtheta}_{u,v}^\top \boldf(\widetilde{x}_i^{(u)P},\widetilde{x}_i^{(v)P})\right)}
{\frac{1}{\widetilde{n}_Q}\sum_{j=1}^{\widetilde{n}_Q} 
\exp\left( \sum_{u'\ge v'} \widehat{\boldtheta}_{u',v'}^\top \boldf(\widetilde{x}_j^{(u')Q},\widetilde{x}_j^{(v')Q}) \right)}.
\end{align*}

\item \textbf{Flasso:}
The sum of feature-wise conditional HOLLs 
for $p(x^{(s)}|\boldx^{(-s)};\boldtheta_s)$ and $q(x^{(s)} | \boldx^{(-s)};\boldtheta_s)$ 
over all nodes is maximized:
\begin{align*}
\frac{1}{\widetilde{n}_P}\sum_{i=1}^{\widetilde{n}_P} \sum_{s=1}^d
 \log p(\widetilde{x}^{(s)}_{i}{}^P | \widetilde{\boldx}^{(-s)}_{i}{}^P;\widehat{\boldtheta}_s^P)
+
\frac{1}{\widetilde{n}_Q}\sum_{i=1}^{\widetilde{n}_Q} \sum_{s=1}^d
 \log q(\widetilde{x}^{(s)}_{i}{}^Q | \widetilde{\boldx}^{(-s)}_{i}{}^Q ;\widehat{\boldtheta}_s^Q).
\end{align*}

\item \textbf{Glasso:}
The sum of HOLLs
for $p(\boldx;\boldtheta)$ and $q(\boldx;\boldtheta)$ is maximized:
\begin{align*}
\frac{1}{\widetilde{n}_P}\sum_{i=1}^{\widetilde{n}_P}
 \log p(\widetilde{\boldx}_i^P;\widehat{\boldtheta}^P)
+
\frac{1}{\widetilde{n}_Q}\sum_{i=1}^{\widetilde{n}_Q}
 \log q(\widetilde{\boldx}_i^Q;\widehat{\boldtheta}^Q).
\end{align*}
\end{itemize}

When $n=100$, KLIEP and Flasso clearly distinguish changed (dashed lines)
and unchanged (solid lines) edges in terms of parameter magnitude.
However, when the sample size is halved to $n=50$, the separation is visually rather unclear
in the case of Flasso. In contrast, the paths of changed and unchanged edges
are still almost disjoint in the case of KLIEP.
The Glasso method performs rather poorly in both cases.
A similar tendency can be observed also in the P-R curve plot:
When the sample size is $n=100$, KLIEP and Flasso work equally well,
but KLIEP gains its lead when the sample size is reduced to $n=50$.
Glasso does not perform well in both cases. 

\setcounter{subfigure}{0}

\subsection{Nonparanormal Distribution}
\label{sec.toy.npn}

\begin{figure*}[t]
\centering
\subfigure[KLIEP, $n = 100$]{
\includegraphics[width = .31\textwidth]{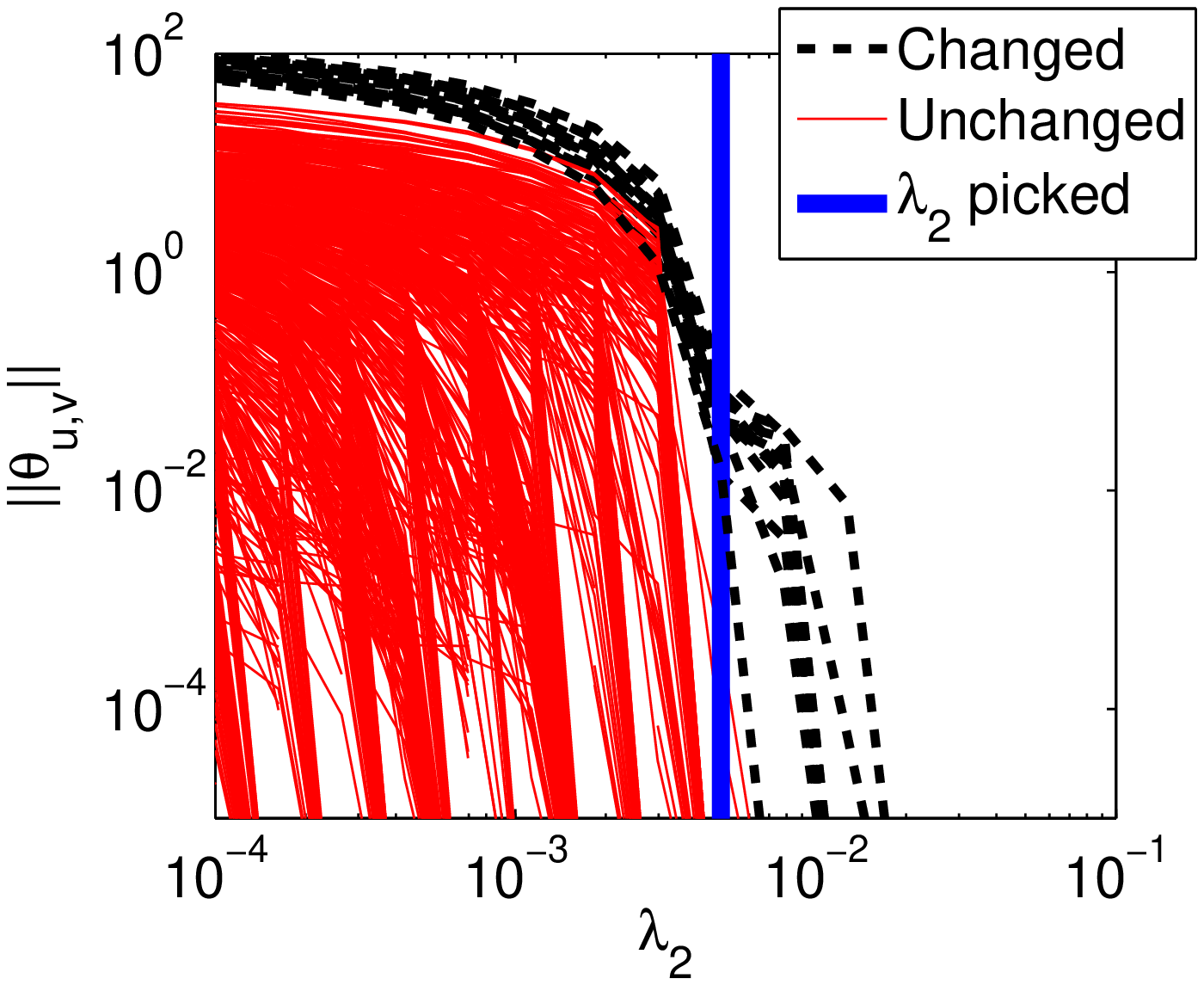}
\label{fig.toy.npn100.kliep}
}
\subfigure[Flasso, $n = 100$]{
\includegraphics[width = .31\textwidth]{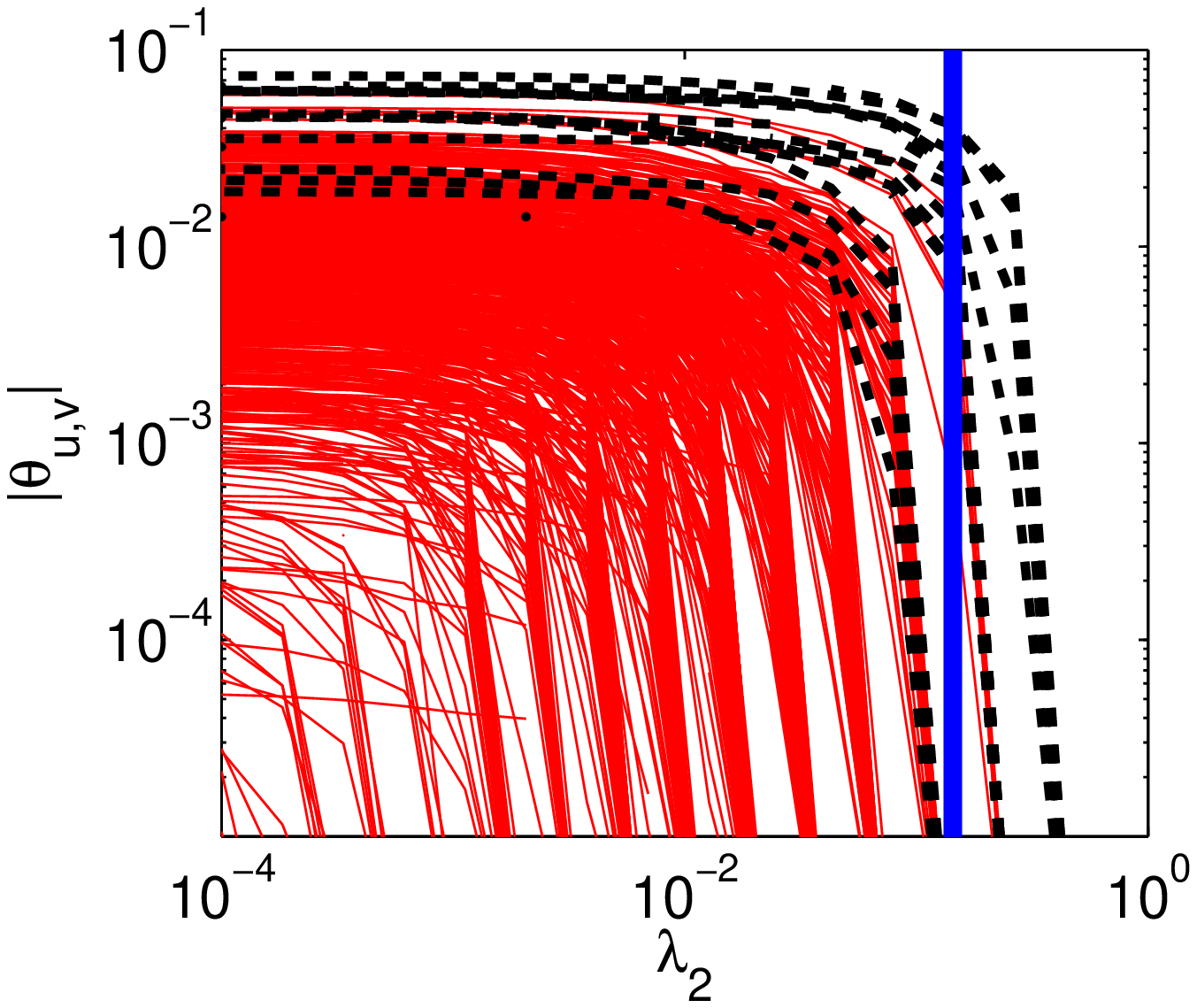}
\label{fig.toy.npn100.fl}
}
\subfigure[Glasso, $n = 100$]{
\includegraphics[width = .31\textwidth]{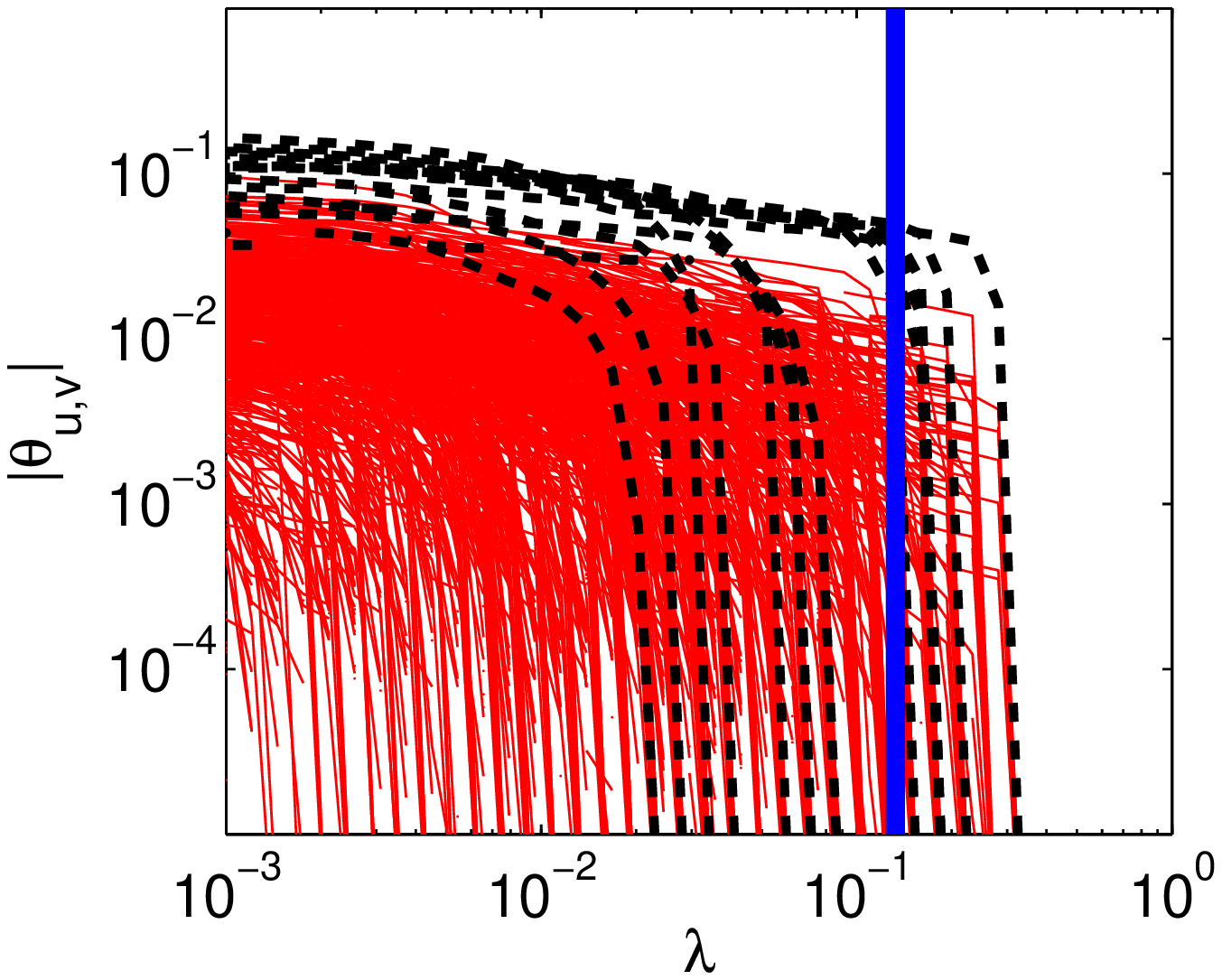}
\label{fig.toy.npn100.gl}
}
\subfigure[KLIEP, $n = 50$]{
\includegraphics[width = .31\textwidth]{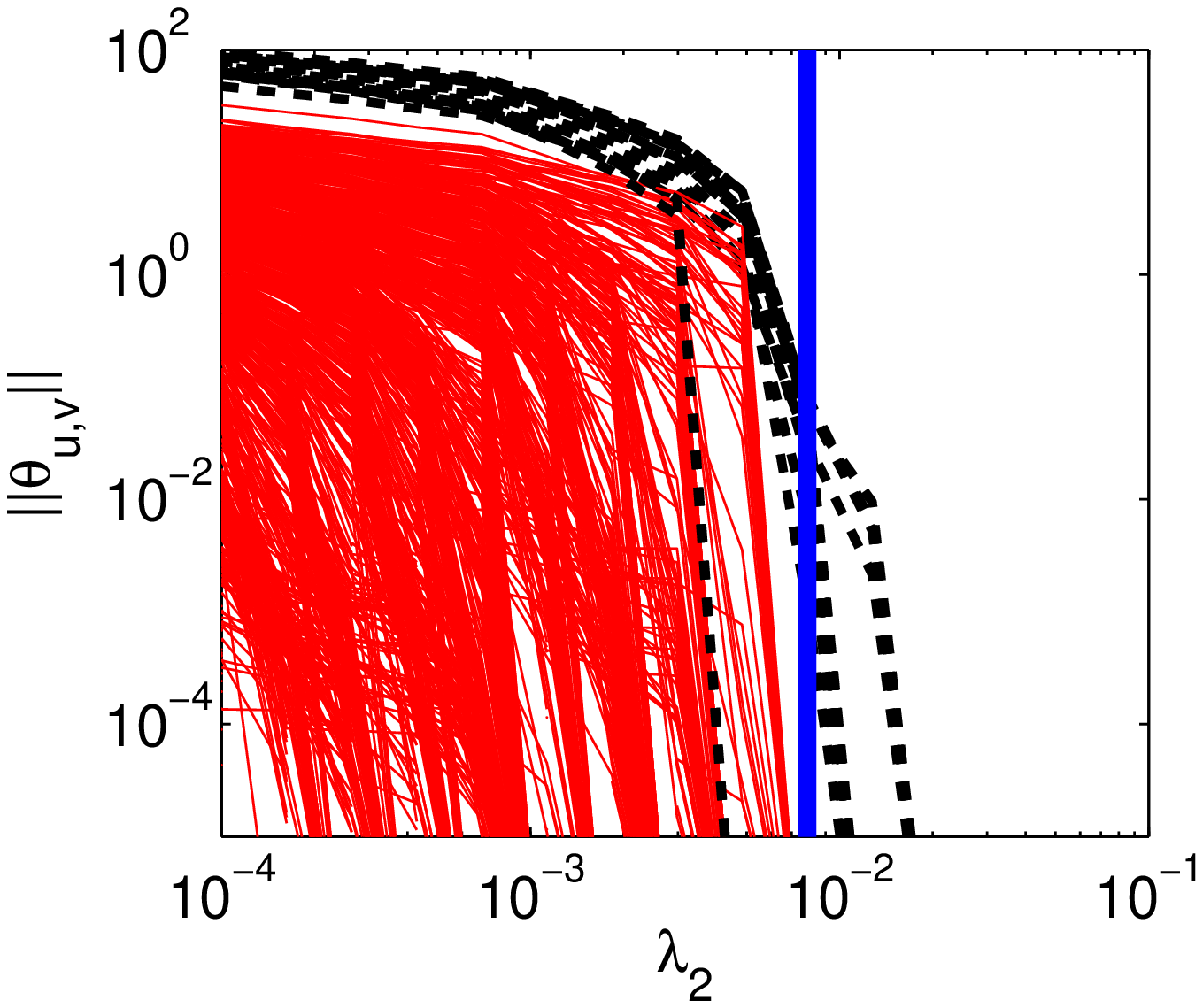}
\label{fig.toy.npn50.kliep}
}
\subfigure[Flasso, $n = 50$]{
\includegraphics[width = .31\textwidth]{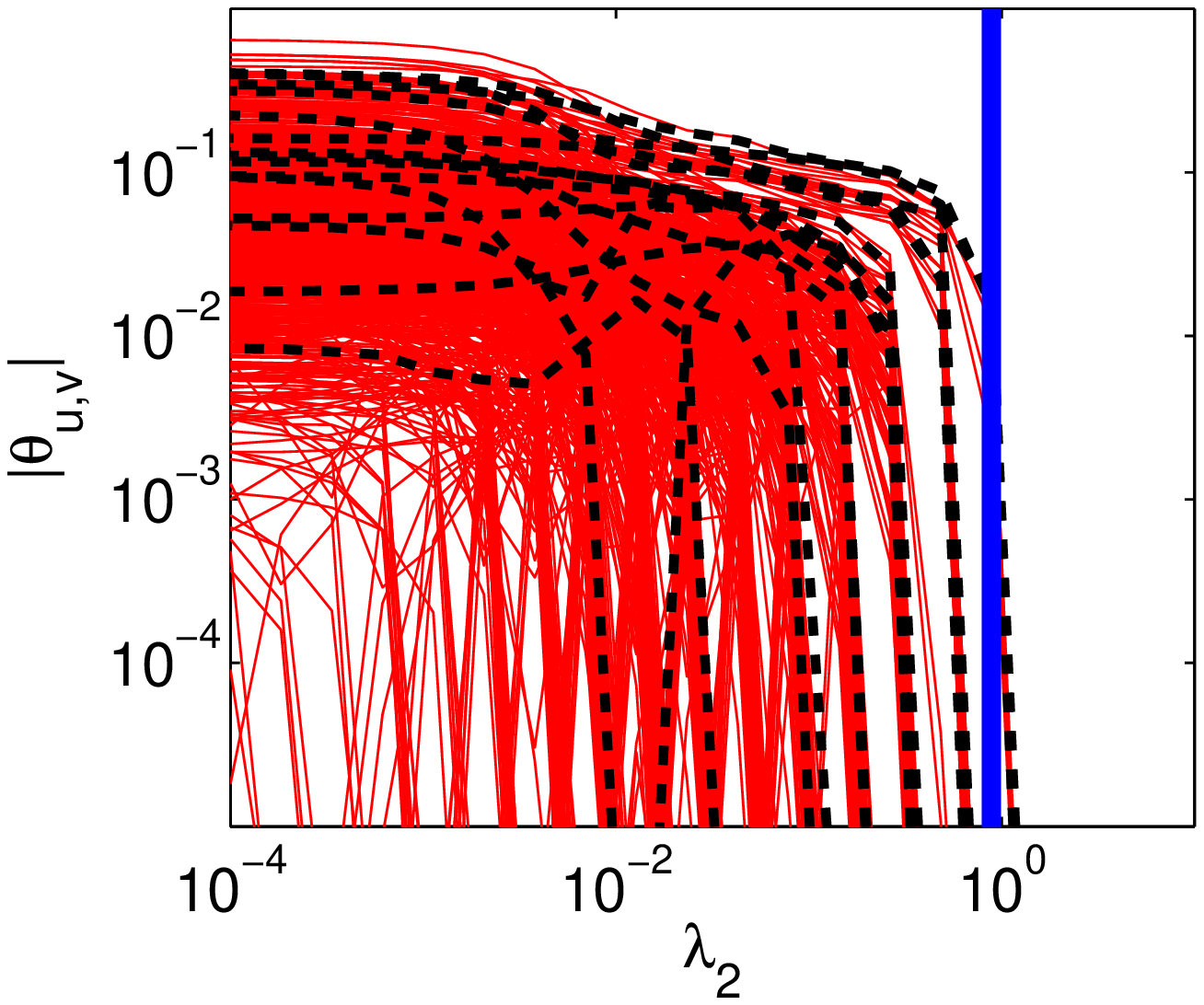}
\label{fig.toy.npn50.fl}
}
\subfigure[Glasso, $n = 50$]{
\includegraphics[width = .31\textwidth]{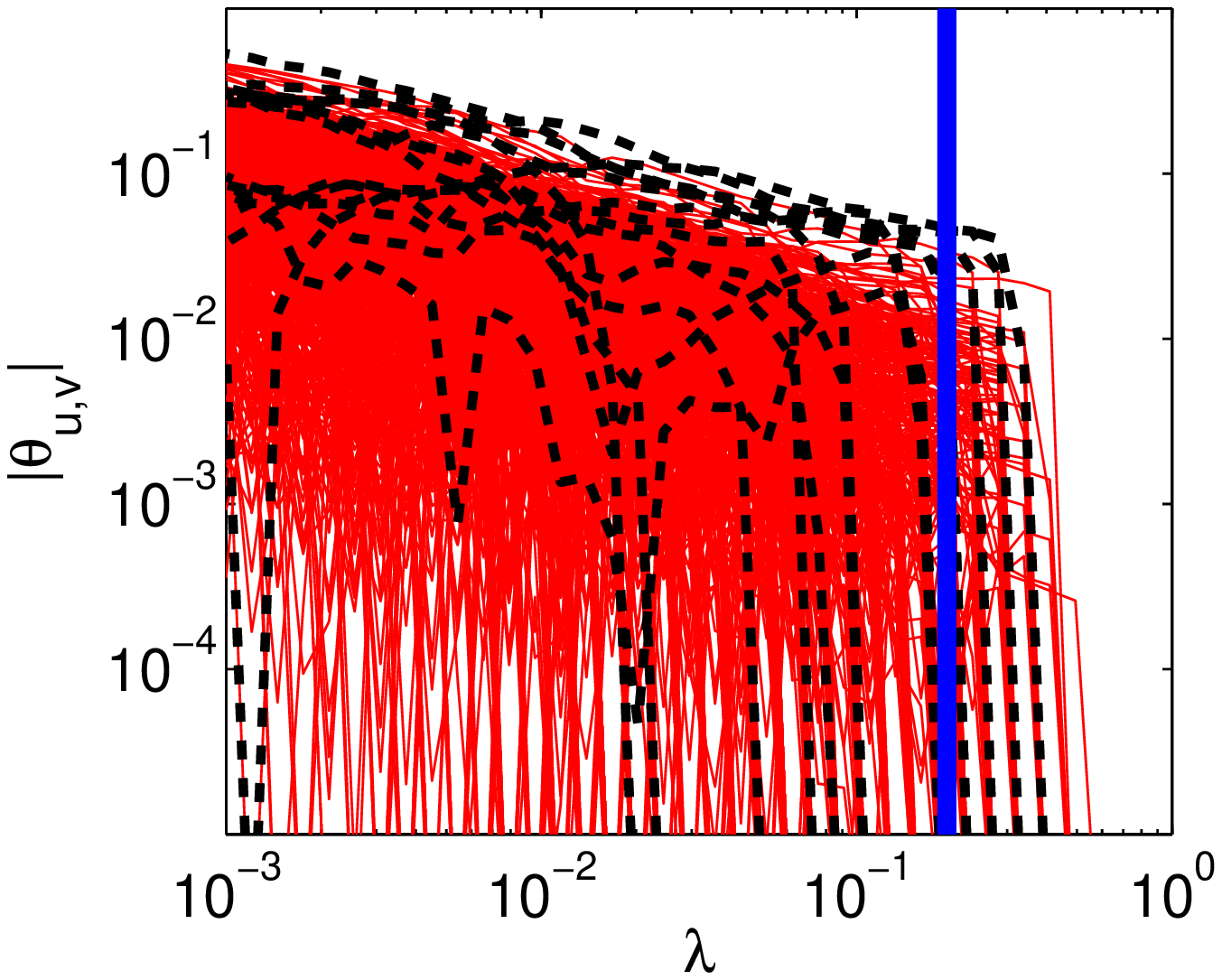}
\label{fig.toy.npn50.gl}
}
\subfigure[Nonparanormal distribution]{
\includegraphics[width = .31\textwidth]{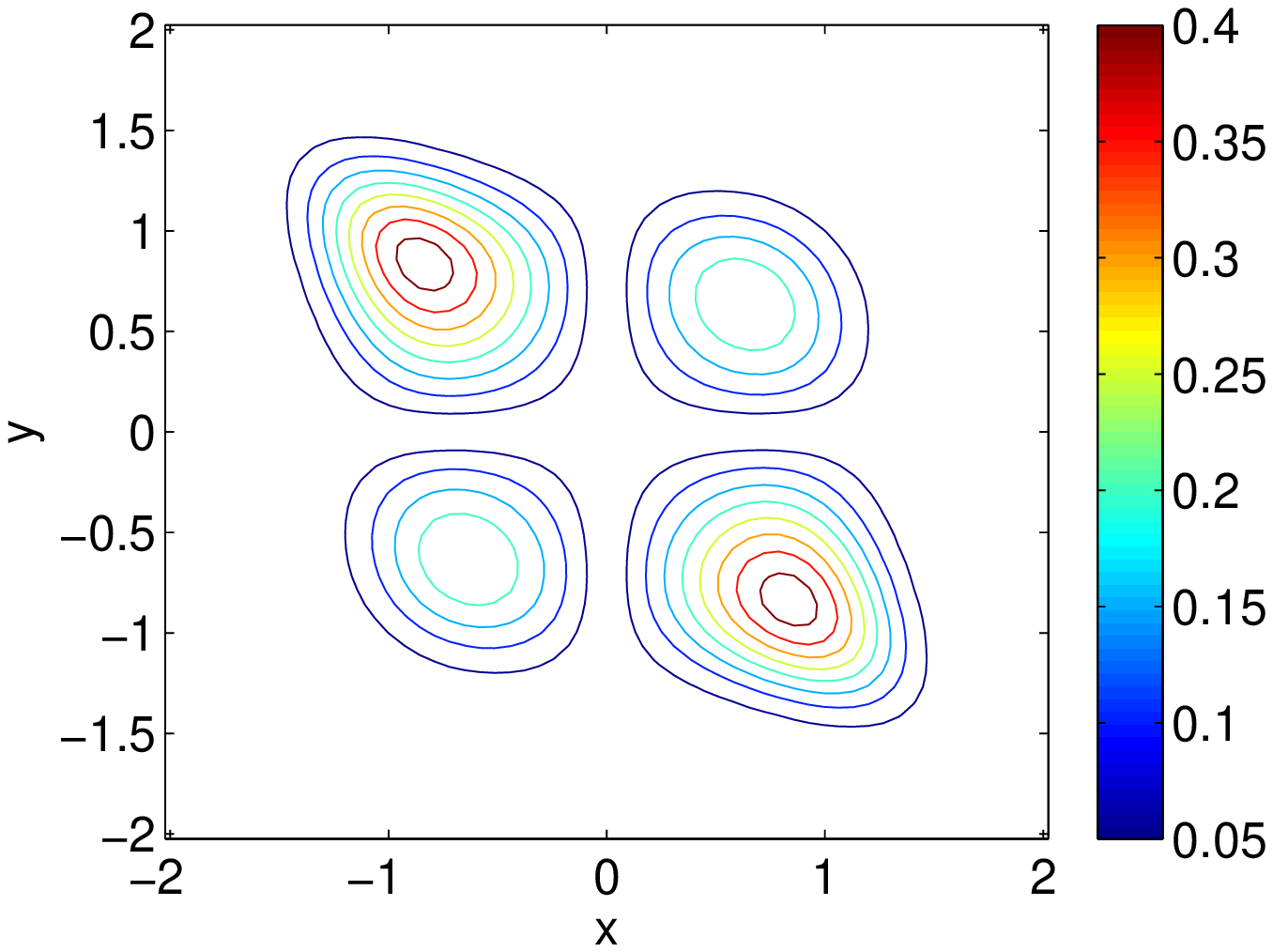}
\label{fig.toy.npn.contour}
}
\subfigure[P-R curve, $n =100$ ]{
\includegraphics[width = .31\textwidth]{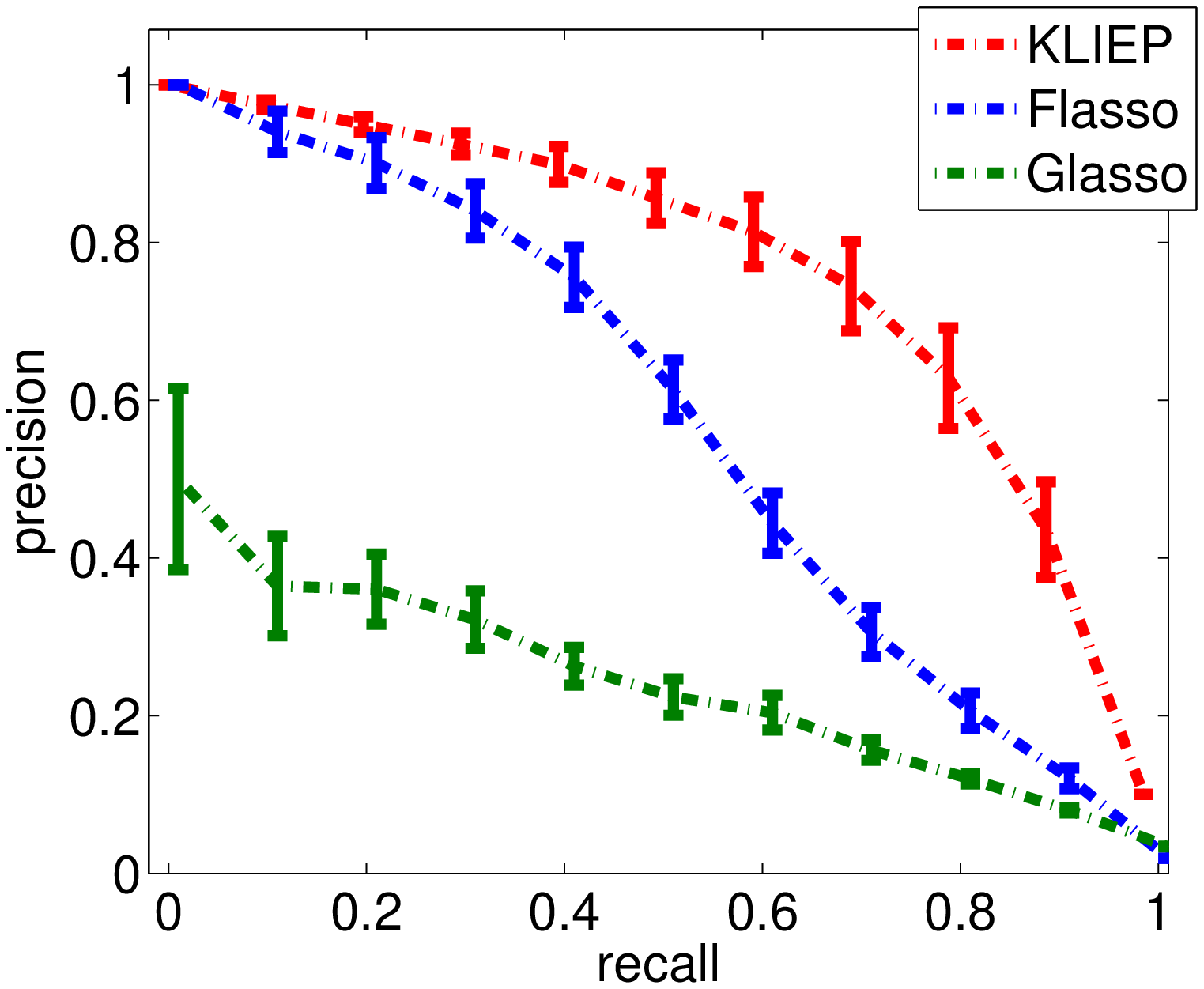}
\label{fig.toy.npn100.pr}
}
\subfigure[P-R curve, $n =50$ ]{
\includegraphics[width = .31\textwidth]{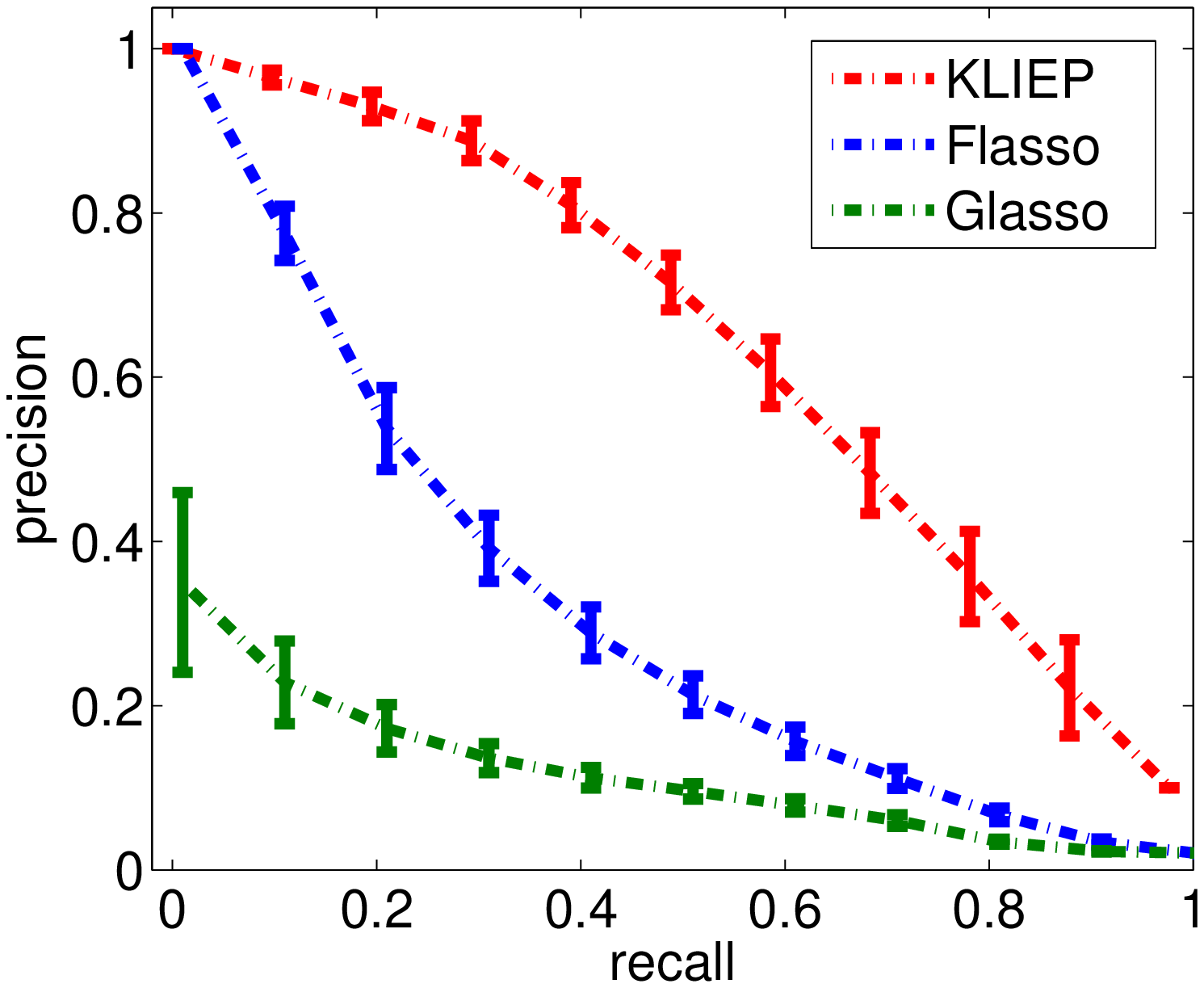}
\label{fig.toy.npn50.pr}
}
\caption{Experimental results on the nonparanormal dataset.}
\label{fig.toy.npn}
\end{figure*}

We post-process the Gaussian dataset used in Section~\ref{sec.toy.gaussian} to construct nonparanormal samples.
More specifically, we apply the power function,
\begin{align*}
h_i^{-1}(x) = \mathrm{sign}(x)|x|^{\frac{1}{2}},
\end{align*}
to each dimension of $\boldx^P$ and $\boldx^Q$,
so that $\boldh(\boldx^P) \sim \mathcal{N}(\boldsymbol{0},({\boldTheta^P})^{-1})$ 
and $\boldh(\boldx^Q) \sim \mathcal{N}(\boldsymbol{0},({\boldTheta^Q})^{-1})$. 

To cope with the non-linearity in the KLIEP method,
we use the power nonparanormal basis functions 
with power $k=2$, $3$, and $4$:
\begin{align*}
\boldf(x_i,x_j)=(\mathrm{sign}(x_{i})|x_{i}|^{k}, \mathrm{sign}(x_{j})|x_{j}|^{k},1)^\top.
\end{align*}
Model selection of $k$ is performed together with the regularization parameter by HOLL maximization.
For Flasso and Glasso, we apply the nonparanormal transform as described in \citet{nonparanormal} before the structural change is learned. 

The experiments are conducted on 20 randomly generated datasets with $n = 50$ and $100$, respectively.
The regularization paths, data generating distribution, and averaged P-R curves are plotted in Figure~\ref{fig.toy.npn}.
The results show that 
Flasso clearly suffers from the performance degradation
compared with the Gaussian case, perhaps because the number of samples is too small for the complicated nonparanormal distribution.
Due to the two-step estimation scheme,
the performance of Glasso is poor.
In contrast, KLIEP separates changed and unchanged edges still clearly
for both $n = 50$ and $n = 100$.
The P-R curves also show the same tendency.

\subsection{``Diamond'' Distribution with No Pearson Correlation}
In the experiments in Section~\ref{sec.toy.npn},
though samples are non-Gaussian, the \emph{Pearson correlation} is not zero.
Therefore, methods assuming Gaussianity can still capture some linear correlation
between random variables. 
Here, we consider a more challenging case
with a diamond-shaped distribution within the exponential family
that has zero Pearson correlation between variables.
Thus, the methods assuming Gaussianity
cannot extract any information in principle from this dataset.

\begin{figure*}
\centering
\setcounter{subfigure}{0}
\subfigure[Diamond distribution]{
\includegraphics[width = .42\textwidth]{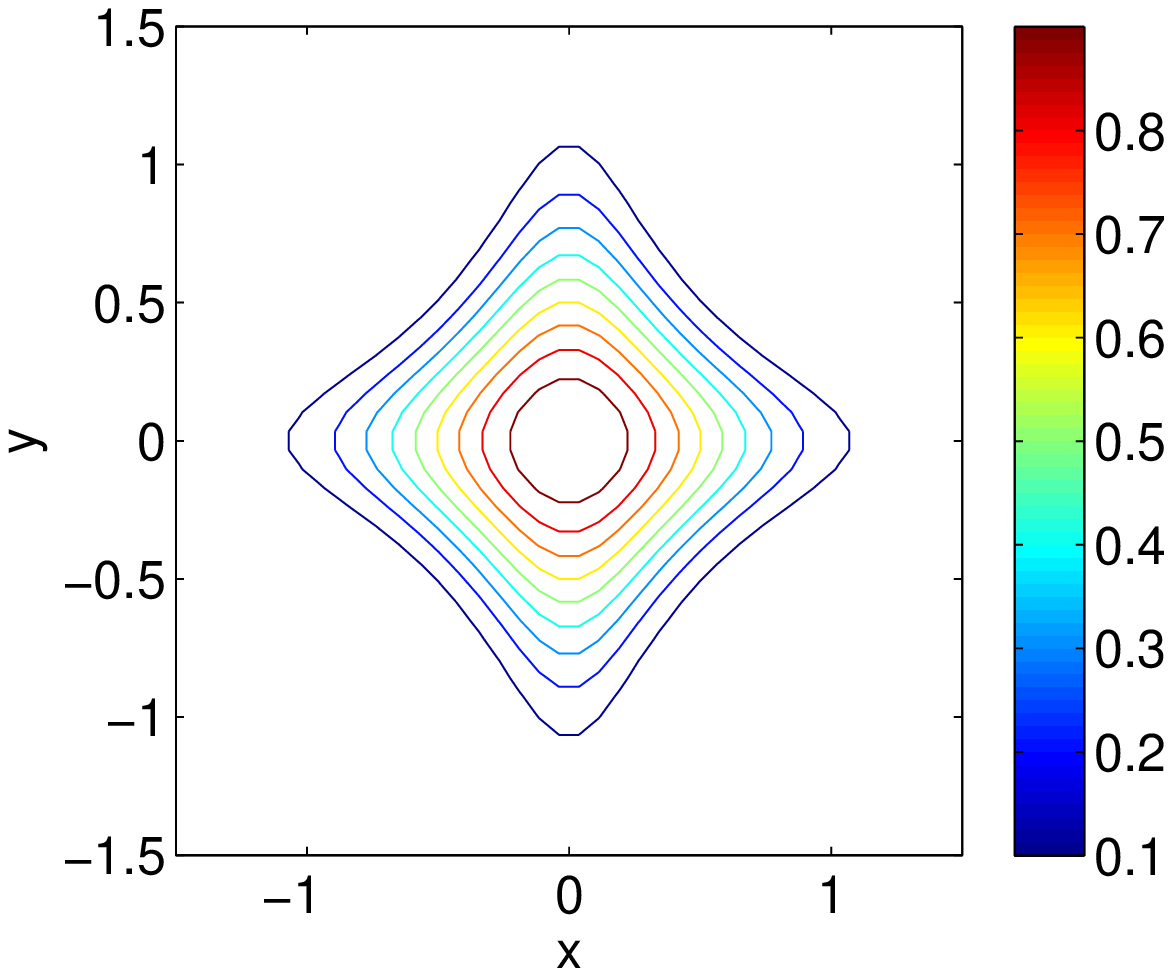}
\label{fig.toy.diamond.contour}
}
\subfigure[KLIEP]{
\label{fig.toy.diamond.kliep}
\includegraphics[width = .42\textwidth]{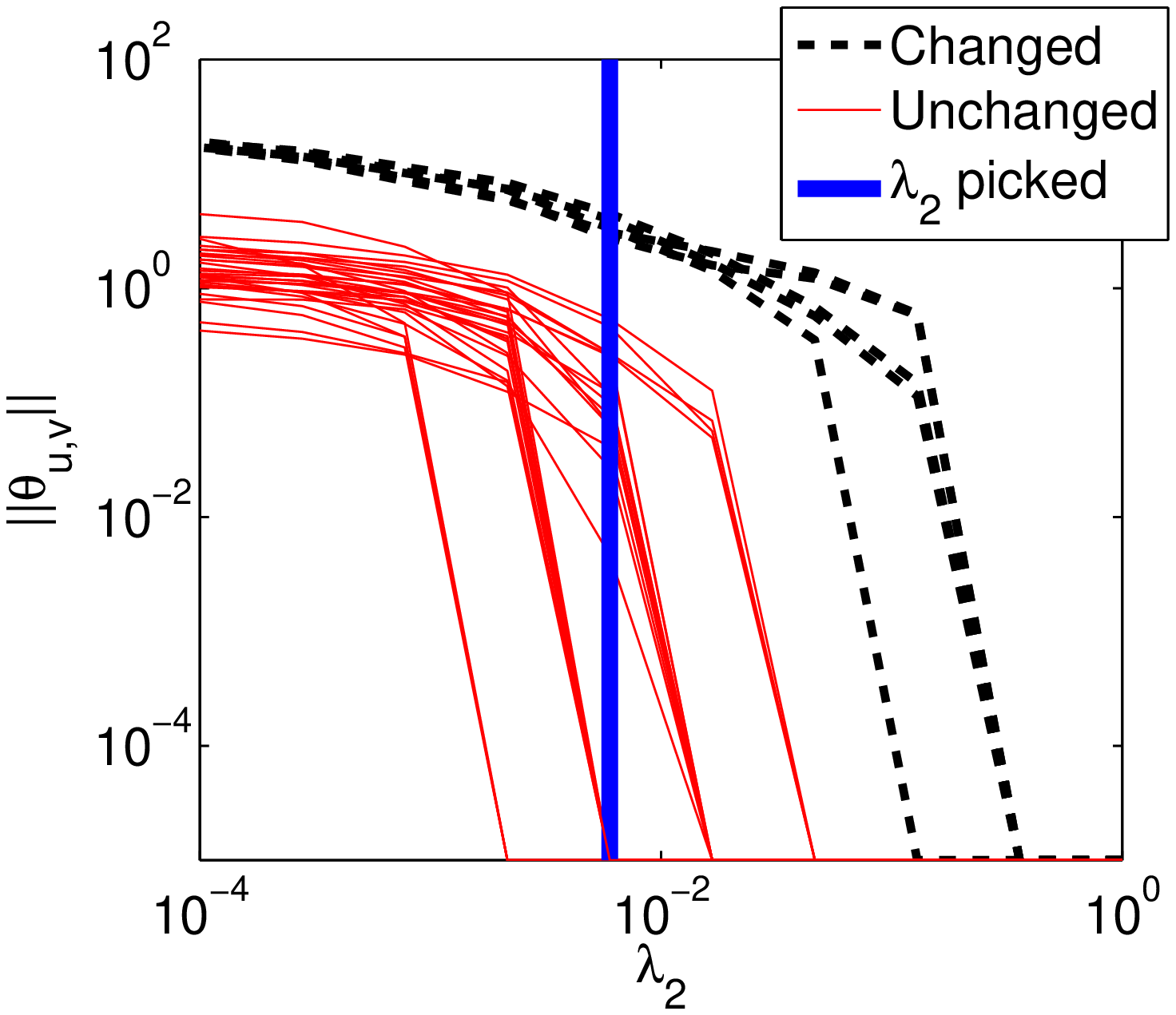}
}\\
\subfigure[IS-Flasso]{
\label{fig.toy.diamond.ismle.fused}
\includegraphics[width = .42\textwidth]{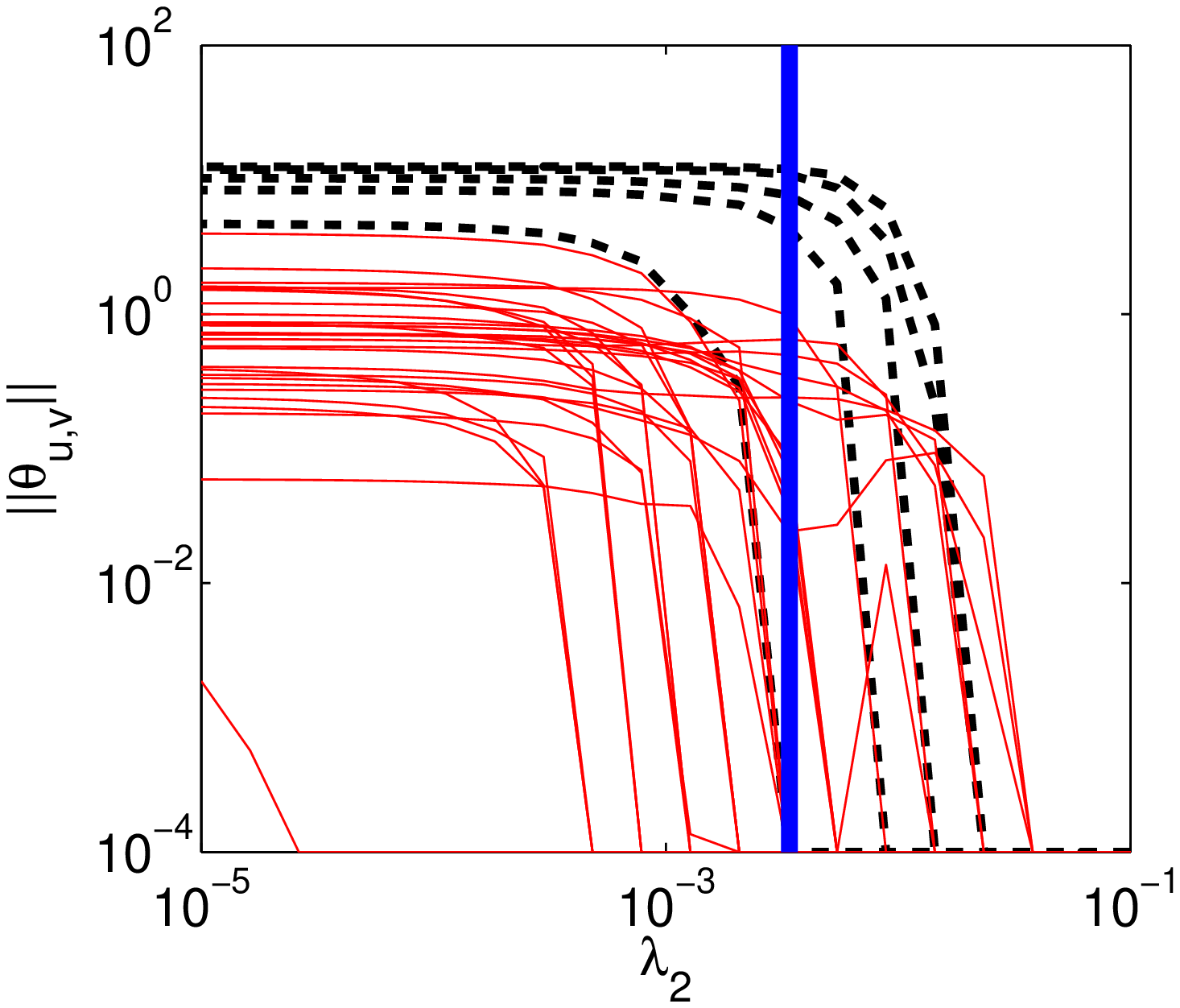}
}
\subfigure[IS-Glasso]{
\label{fig.toy.diamond.ismle}
\includegraphics[width = .42\textwidth]{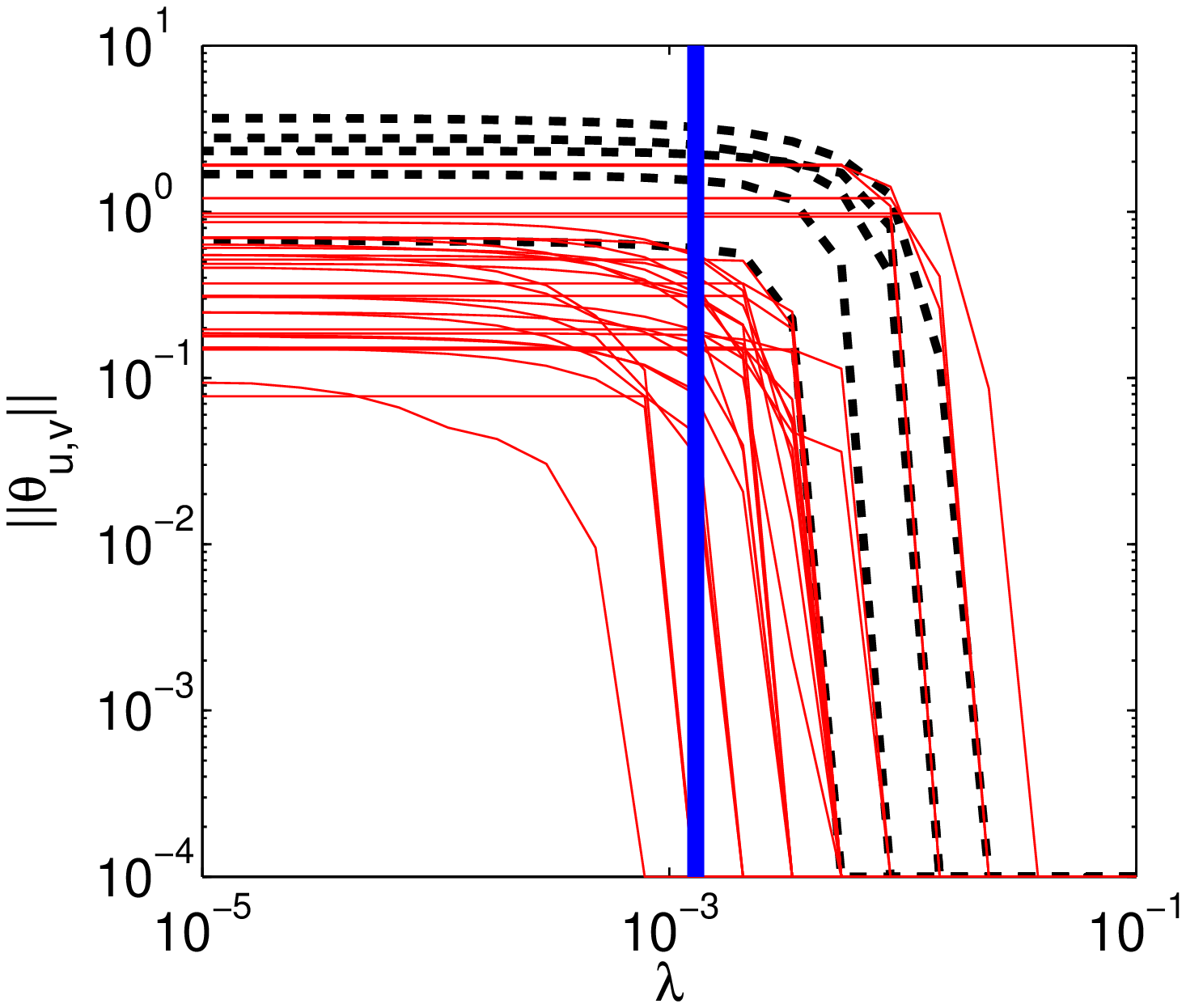}
}
\subfigure[P-R curve]{
\label{fig.toy.diamond.pr}
\includegraphics[width = .8\textwidth]{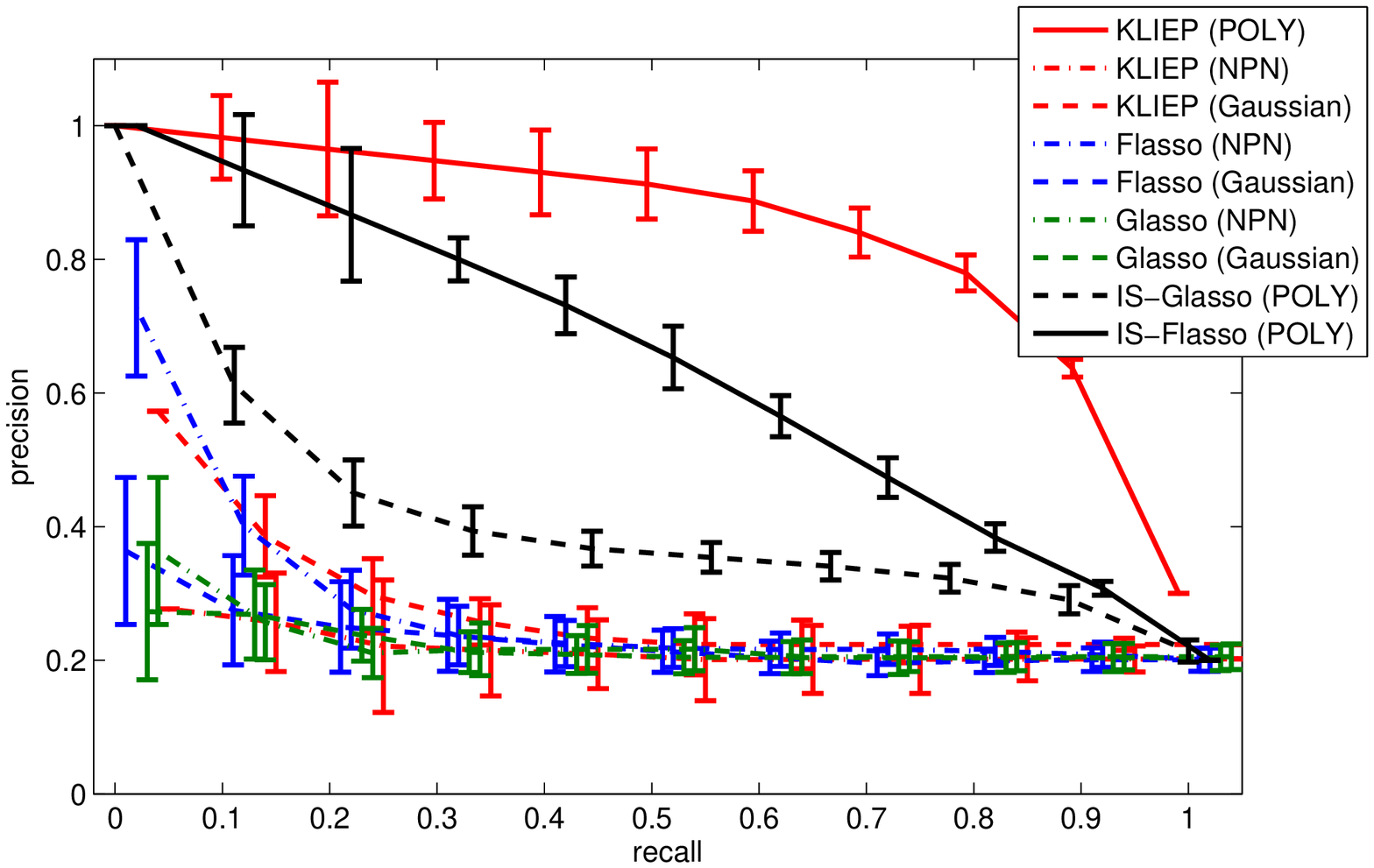}
}
\caption{Experimental results on the diamond dataset.
  ``NPN'' and ``POLY'' denote the nonparanormal and polynomial models, respectively.
  Note that the precision rate of $100\%$ recall for a random guess is approximately $20\%$.}
\label{fig.toy.diamond}
\end{figure*}

The probability density function of the diamond distribution is defined as
follows (Figure~\ref{fig.toy.diamond.contour}):
\begin{align}
p(\boldx) \propto \mathrm{exp}\left(-\sum_{i=1}^d 2x_i^2 - \sum_{(i,j):A_{i,j} \neq 0} 20x_i^2 x_j^2  \right),
\label{eq.diamond}
\end{align}
where the adjacency matrix $\boldA$ describes the MN structure.
Note that this distribution cannot be transformed into a Gaussian distribution by any nonparanormal transformations. 

We set $d = 9$ and  $n_P = n_Q = 5000$.
$\boldA^P$ is randomly generated with $35\%$ sparsity,
while $\boldA^Q$ is created by randomly removing edges in $\boldA^P$
so that the sparsity level is dropped to $15\%$.
Samples from the above distribution are drawn by using a \emph{slice sampling} method \citep{slicesampling}.
Since generating samples from high-dimensional distributions is non-trivial and time-consuming,
we focus on a relatively low-dimensional case.
To avoid sampling error which may mislead the experimental evaluation, we also increase the sample size, so that the erratic points generated by accident will not affect the overall population.

In this experiment, we compare the performance of KLIEP, Flasso, and Glasso
with the Gaussian model, the power nonparanormal model,
and the polynomial model:
\begin{align*}
\boldf(x_i,x_j)=(x_i^k, x_j^k, x_ix_j^{k-1}, \dots, x_i^{k-1}x_j,x_i^{k-1},x_j^{k-1},\dots,x_i,x_j,1)^\top \text{ for } i \neq j. 
\end{align*}
The univariate polynomial transform is defined as $\boldf(x_i,x_i) = \boldf(x_i, 0)$. 
We test $k=2,3,4$ and choose the best one in terms of HOLL.
The Flasso and Glasso methods for the polynomial model
are computed by importance sampling, i.e., we use the IS-Flasso and IS-Glasso methods
(see Section~\ref{sec.ismle}).
Since these methods are computationally very expensive, we only test $k=4$
which we found to be a reasonable choice.
We set the instrumental distribution $p'$ as the standard normal
$\mathcal{N}(\boldsymbol{0}, \boldI)$, 
and use sample $\{\boldx'_i\}_{i=1}^{70000}\sim p'$ for approximating integrals.
$p'$ is purposely chosen so that 
it has a similar ``bell'' shape to the target densities
but with larger variance on each dimension.

The averaged P-R curves over 20 datasets are shown in Figure~\ref{fig.toy.diamond.pr}. 
KLIEP with the polynomial model significantly outperforms all the other methods, 
while the IS-Glasso and especially IS-Flasso give better result than the KLIEP,
Flasso, and Glasso methods with the Gaussian and nonparanormal models.
This means that the polynomial basis function is indeed helpful
in handling completely non-Gaussian data.
However, as discussed in Section~\ref{sec.MLE},
it is difficult to use such a basis function
in Glasso and Flasso
because of the computational intractability of the normalization term. 
Although IS-Glasso can approximate integrals, 
the result shows that such approximation of integrals
does not lead to a very good performance. In comparison, the result of the IS-Flasso method is much improved thanks to the coupled sparsity regularization, but it is still not comparable to KLIEP.


The regularization paths of KLIEP with the polynomial model
illustrated in Figure~\ref{fig.toy.diamond.kliep}
show the usefulness of the proposed method in change detection under non-Gaussianity. 
We also give regularization paths obtained by the IS-Flasso and IS-Glasso methods on the same dataset
in Figures~\ref{fig.toy.diamond.ismle.fused} and \ref{fig.toy.diamond.ismle}, respectively.
The graphs show that both methods do not separate changed and unchanged edges well,
though the IS-Flasso method works slightly better. 



\subsection{Computation Time: Dual versus Primal Optimization Problems}
\begin{figure*}[t]
\centering
\includegraphics[width = .5\textwidth]{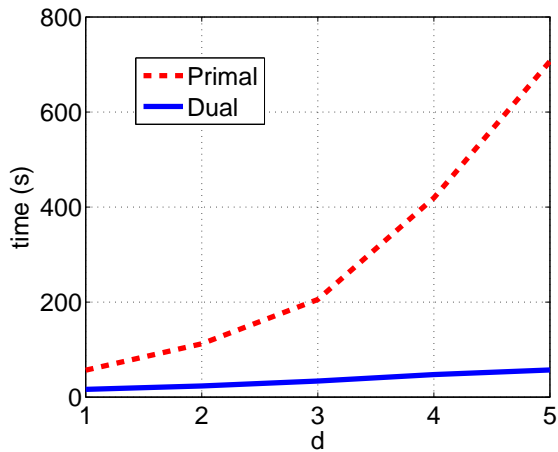}
\caption{Comparison of computation time for solving primal and dual optimization problems. }
\label{fig.dualvsprimal}
\end{figure*}

Finally, we compare the computation time of the proposed KLIEP method
when solving the dual optimization problem \eqref{eq.obj.dual} and the primal optimization problem \eqref{eq.final.obj}. Both the optimization problems are solved by using the same convex optimizer \emph{minFunc}\footnote{\url{http://www.di.ens.fr/~mschmidt/Software/minFunc.html}}.
The datasets are generated from two Gaussian distributions constructed in the same way as Section~\ref{sec.toy.gaussian}.
150 samples are separately drawn from two distributions with dimension $d = 40, 50, 60, 70, 80$.
We then perform change detection by computing the regularization paths using 20 choices of $\lambda_2$ ranging from $10^{-4}$ to $10^0$ and fix $\lambda_1 = 0.1$.
The results are plotted in Figure~\ref{fig.dualvsprimal}.

It can be seen from the graph that as the dimensionality increases, the computation time for solving the primal optimization problem is sharply increased, while that for solving the dual optimization problem grows only moderately: when $d=80$, the computation time for obtaining the primal solution is almost 10 times more than that required for obtaining the dual solution. 
Thus, the dual formulation is computationally much more efficient than the primal formulation.

\section{Applications}
In this section, we report the experimental results
on a synthetic gene expression dataset and a Twitter dataset.

\label{sec.real.world}
\subsection{Synthetic Gene Expression Dataset}
\label{sec.sygene}
A gene regulatory network encodes interactions between DNA segments.
However, the way genes interact may change due to environmental or biological stimuli.
In this experiment, we focus on detecting such changes.
We use \emph{SynTReN},
which is a generator of gene regulatory networks
used for benchmark validation of bioinformatics algorithms \citep{Van_den_Bulcke_SynTReN}. 

\begin{figure*}
\centering
\subfigure[Gene regulatory network]{
\includegraphics[clip, width=.6\textwidth]{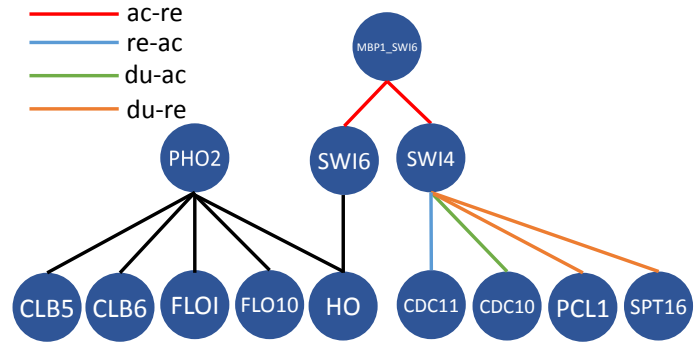}
\label{fig.gene_graph}
}\\
\subfigure[KLIEP]{\includegraphics[width=.4\textwidth]{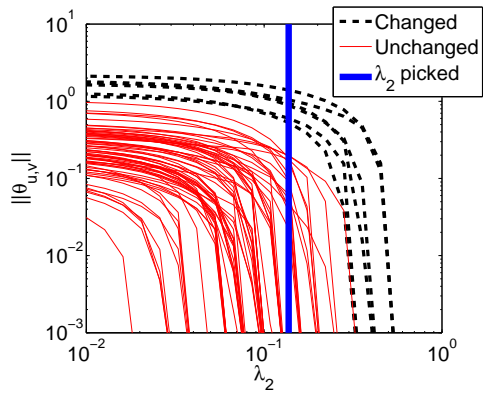}
\label{fig.gene_csl}
}
\subfigure[IS-Flasso]{\includegraphics[width=.4\textwidth]{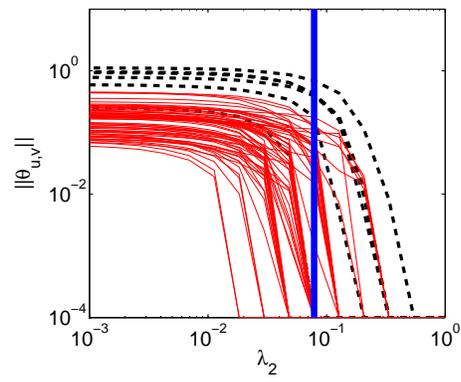}
\label{fig.gene_ISFL}
}
\subfigure[Flasso]{\includegraphics[width=.4\textwidth]{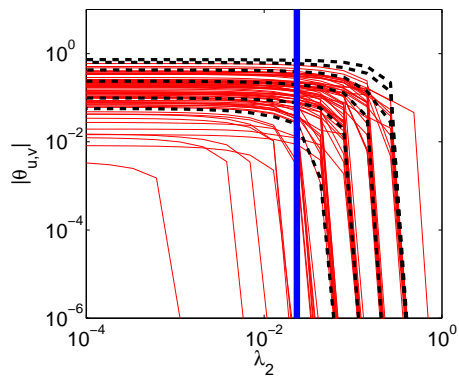}
\label{fig.gene_gaussian}
}
\subfigure[P-R curve]{\includegraphics[width=.4\textwidth]{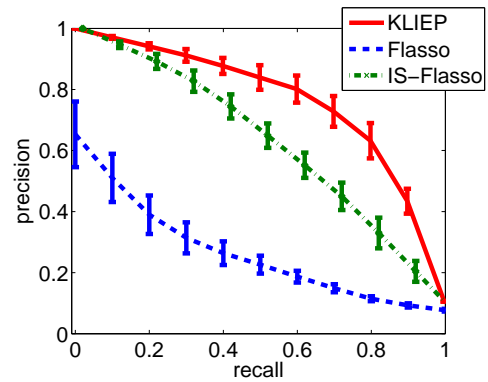}
\label{fig.pr.gene}
}

\caption{Experiments on synthetic gene expression datasets.}
\end{figure*}

We first choose a sub-network containing $13$ nodes from an existing signaling network
in \emph{Saccharomyces cerevisiae} (shown in Figure~\ref{fig.gene_graph}).
Three types of interactions are modeled:
activation (ac), deactivation (re), and dual (du).
$50$ samples are generated in the first stage,
after which we change the types of interactions in $6$ edges,
and generate $50$ samples again.
Four types of changes are considered: 
ac $\rightarrow$ re, re $\rightarrow$ ac, du $\rightarrow$ ac, and du $\rightarrow$ re.

We use KLIEP and IS-Flasso with the polynomial transform function for $k \in \{2,3,4\}$.
The regularization parameter $\lambda_1$ in KLIEP and Flasso 
is tested with choices $\lambda_1 \in \{0.1, 1, 10\}$.
We set the instrumental distribution $p'$ as the standard normal 
$\mathcal{N}(\boldsymbol{0}, \boldI)$, 
and use sample $\{\boldx'_i\}_{i=1}^{70000}\sim p'$ for approximating integrals
in IS-Flasso.



The regularization paths on one example dataset for KLIEP, IS-Flasso,
and the plain Flasso with the Gaussian model are plotted in 
Figures~\ref{fig.gene_csl}, \ref{fig.gene_ISFL}, and \ref{fig.gene_gaussian}, respectively.
Averaged P-R curves over $20$ simulation runs are shown in Figure~\ref{fig.pr.gene}. 
We can see clearly from the KLIEP regularization paths shown in Figure~\ref{fig.gene_csl}
that
the magnitude of estimated parameters on the changed pairwise interactions is much higher than that of the unchanged edges.
IS-Flasso also achieves rather clear separation between changed and unchanged interactions,
though there are a few unchanged interactions drop to zero at the final stage. 
Flasso gives many false alarms by assigning non-zero values
to the unchanged edges, even after some changed edges hit zeros. 

Reflecting a similar pattern, the P-R curves plotted in Figure~\ref{fig.pr.gene}
show that the proposed KLIEP method has the best performance among all three methods.
We can also see that
the IS-Flasso method achieves significant improvement over the plain Flasso method
with the Gaussian model. 
The improvement from Flasso to IS-Flasso shows that the use of the polynomial
basis is useful on this dataset, and the
improvement from IS-Flasso to KLIEP shows that the direct estimation
can further boost the performance.


\subsection{Twitter Story Telling}
\label{sec.twitter}
Finally, we use KLIEP with the polynomial transform function for $k \in \{ 2,3,4\}$ and Flasso as event detectors
from Twitter.
More specifically, we choose the \emph{Deepwater Horizon oil spill}\footnote{\url{http://en.wikipedia.org/wiki/Deepwater_Horizon_oil_spill}} 
as the target event,
and we hope that our method can recover some story lines from Twitter as the news events develop.
Counting the frequencies of $10$ keywords
(BP, oil, spill, Mexico, gulf, coast, Hayward, Halliburton, Transocean, and Obama),
we obtain a dataset by sampling $4$ times per day from February 1st, 2010 to October 15th, 2010,
resulting in $1061$ data samples.


We segment the data into two parts:
the first $300$ samples collected before the day of oil spill (April 20th, 2010) 
are regarded as conforming to a $10$-dimensional joint distribution $Q$,
while the second set of samples that are in an arbitrary $50$-day window
after the oil spill accident happened is regarded as following distribution $P$. 
Thus, the MN of $Q$ encodes the original conditional independence
of frequencies between $10$ keywords,
while the underlying MN of $P$ has changed since an event occurred.
We expect that unveiling changes in MNs between $P$ and $Q$ can recover 
the drift of popular topic trends
on Twitter in terms of the dependency among keywords.

\begin{figure}[t]
\centering
\subfigure[April 17th--June 5th, KLIEP]{\label{fig.twitter.proposed.april}\includegraphics[width=.31\textwidth]{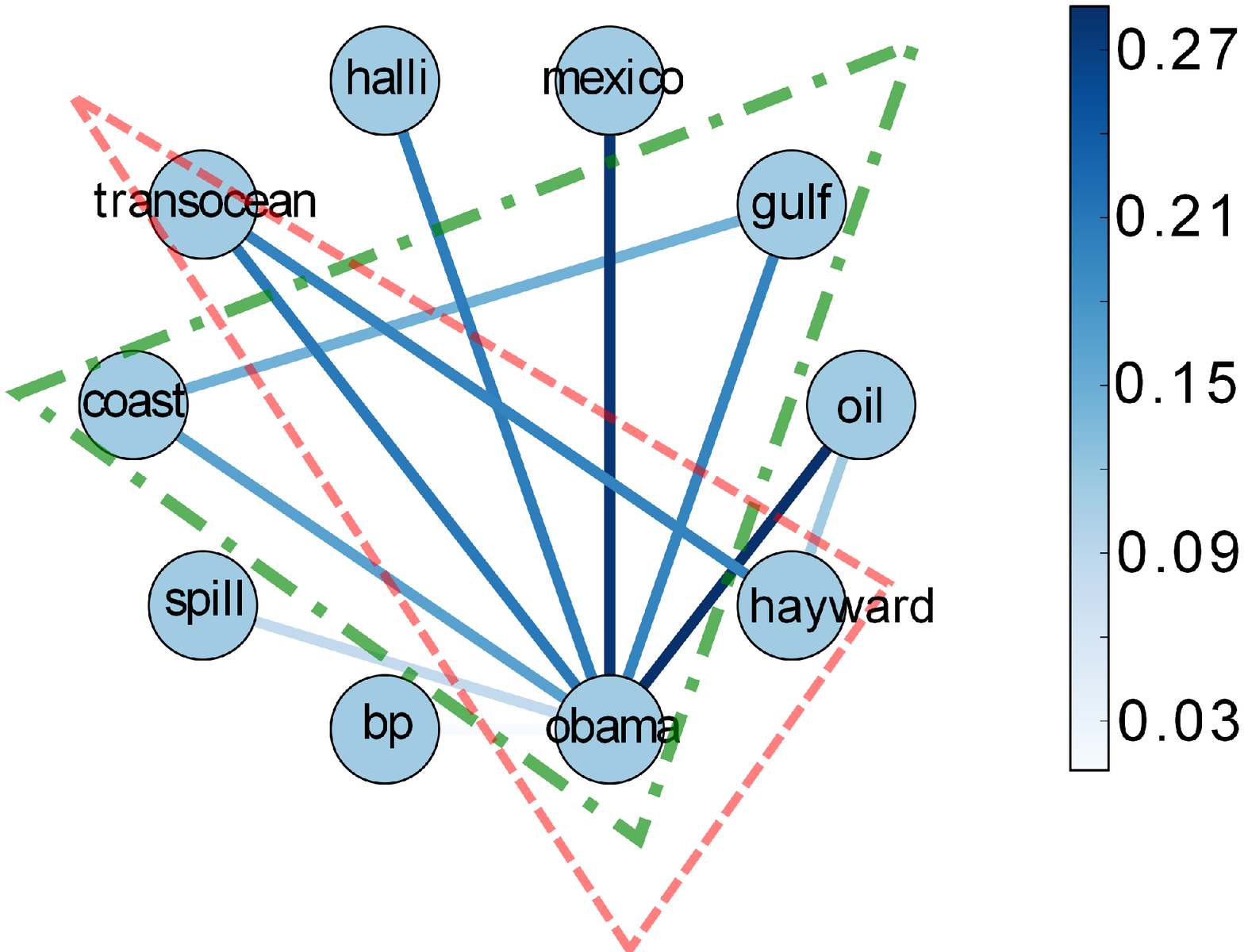}}
\subfigure[June 6th--July 25th, KLIEP]{\label{fig.twitter.proposed.june}\includegraphics[width=.31\textwidth]{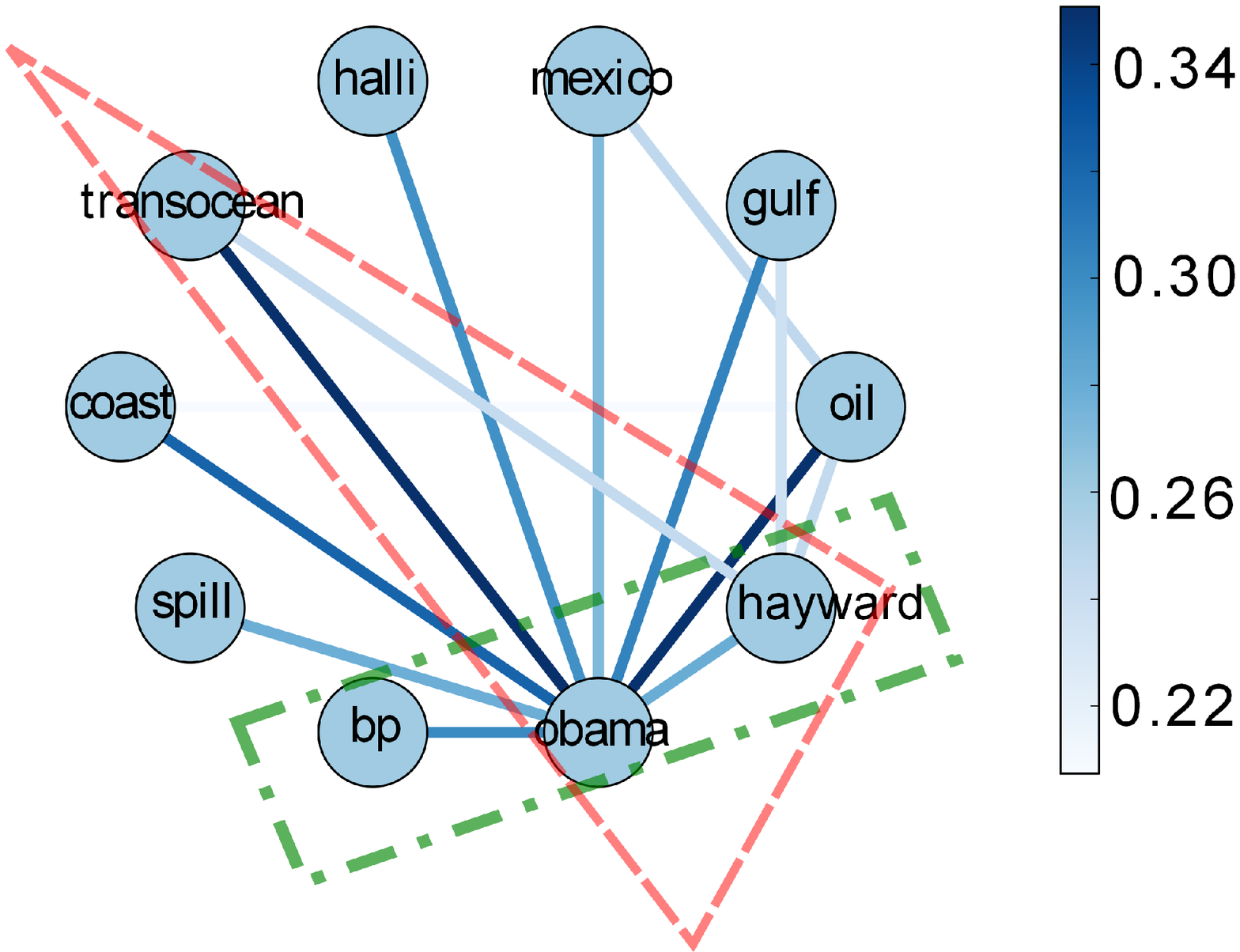}}
\subfigure[July 26th--Sept. 14th, KLIEP]{
\label{fig.twitter.proposed.july}\includegraphics*[width=.31\textwidth]{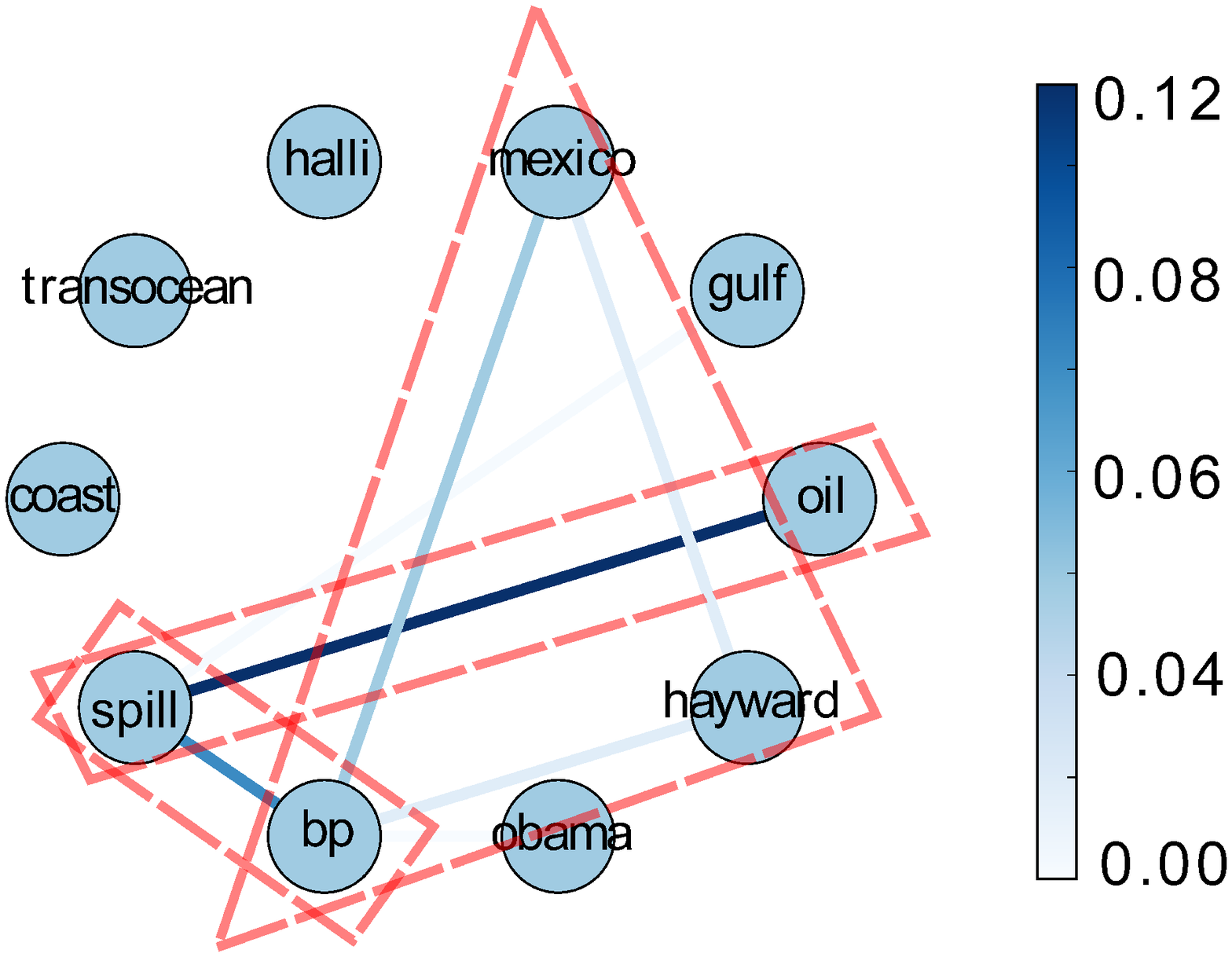}}\\
\subfigure[April 17th--June 5th, Flasso]{\label{fig.twitter.fl.april}\includegraphics*[width=.31\textwidth]{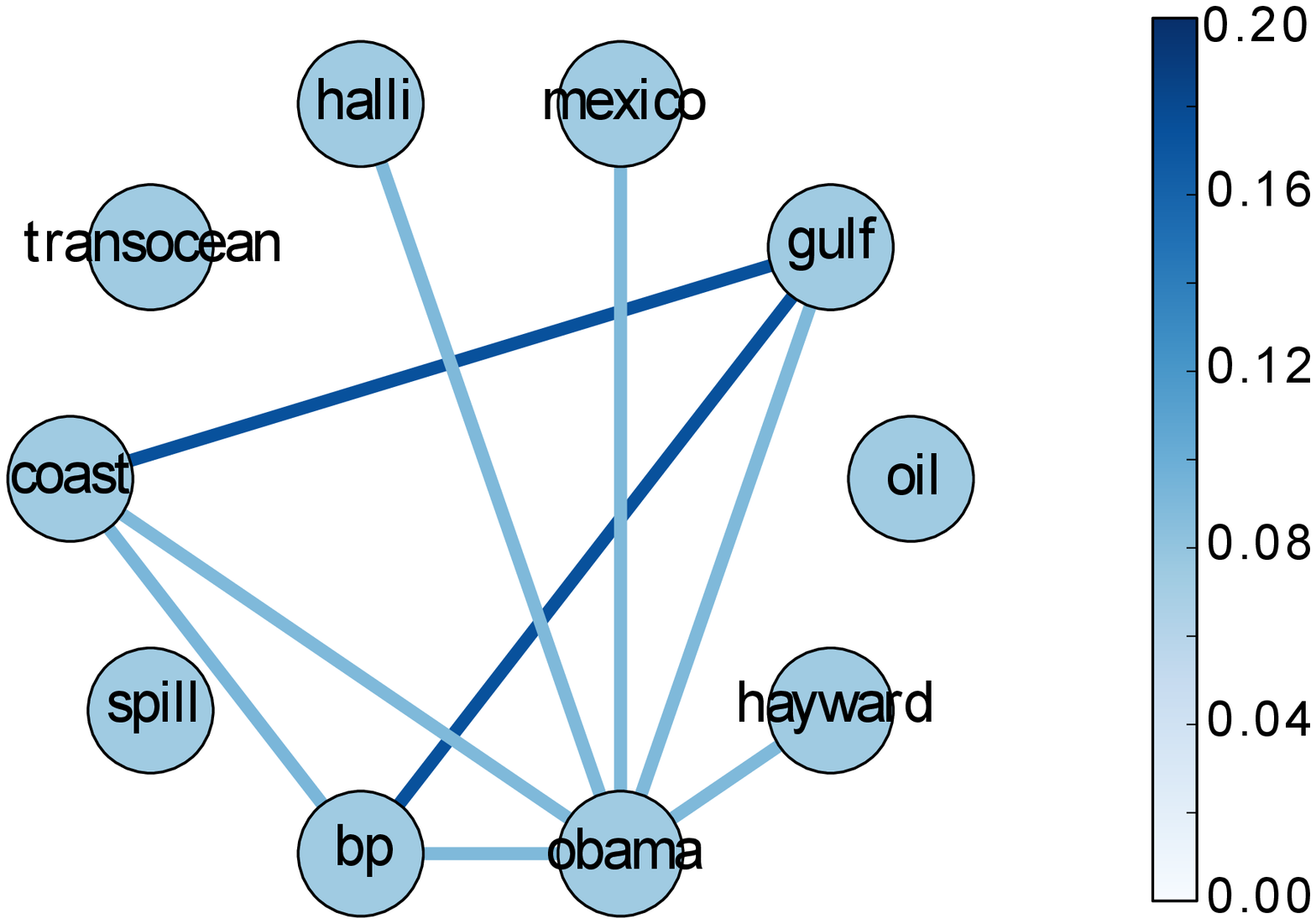}}~~
\subfigure[June 6th--July 25th, Flasso]{\label{fig.twitter.fl.june}\includegraphics*[width=.31\textwidth]{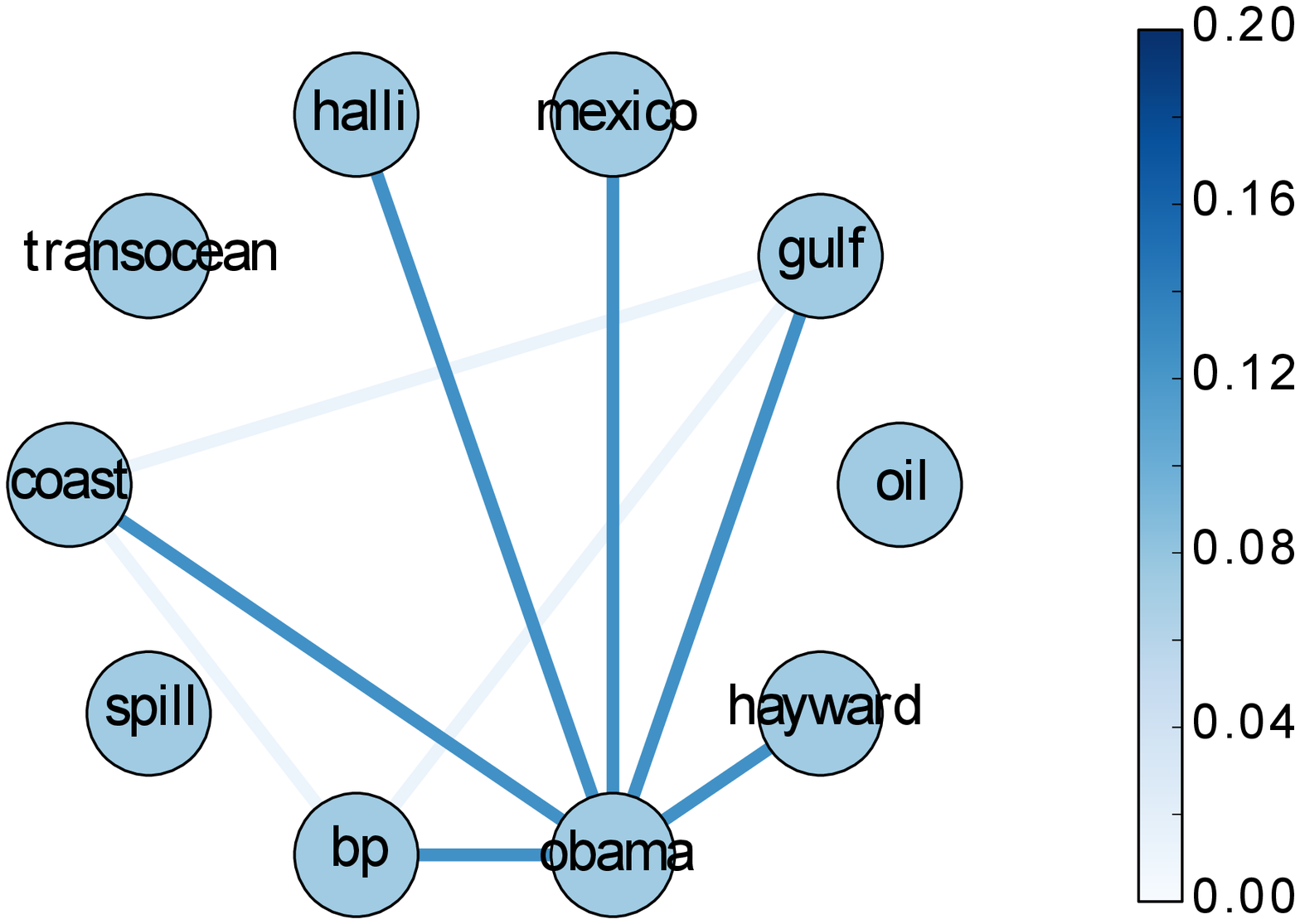}}~~
\subfigure[July 26th--Sept. 14th, Flasso]{
\label{fig.twitter.fl.july}\includegraphics[width=.25\textwidth]{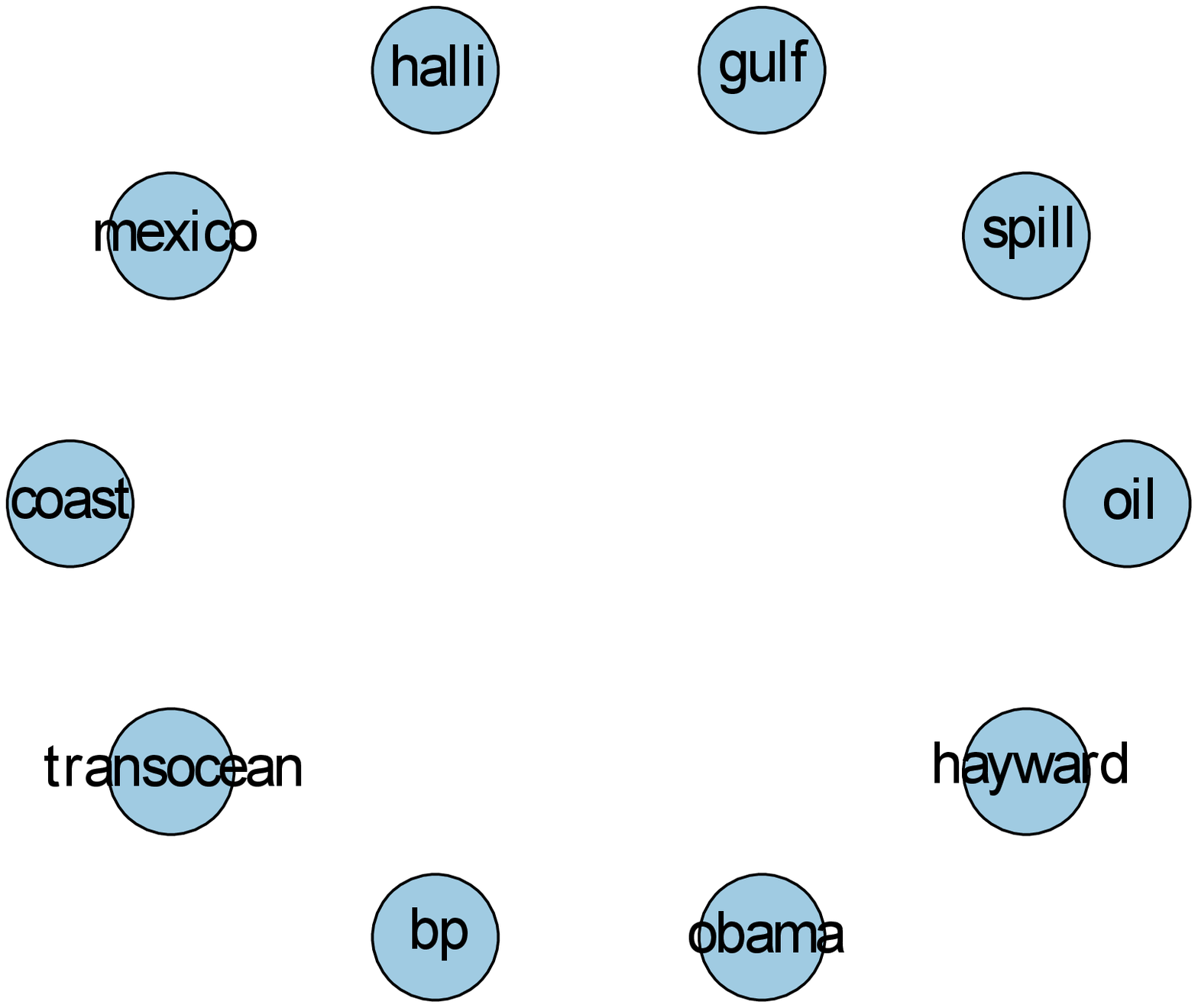}}
\caption{Change graphs captured by the proposed KLIEP method (top) and the Flasso method (bottom).
  The date range beneath each figure indicates when $P$ was sampled, while $Q$ is fixed to dates from February 1st to April 20th.
  Notable structures shared by the graph of both methods are surrounded by the dash-dotted lines. 
  Unique structures that only appear in the graph of the proposed KLIEP method are surrounded by
  the dashed lines.}
\label{fig.twitter}
\end{figure}

The detected change graphs (i.e., the graphs with only detected changing edges) on $10$ keywords are illustrated in Figure~\ref{fig.twitter}.
The edges are selected at a certain value of $\lambda_2$
indicated by the maximal \emph{cross-validated log-likelihood} (CVLL). 
Since the edge set that is picked by CVLL
may not be sparse in general, we sparsify the graph based on the permutation test
as follows:
we randomly shuffle the samples between $P$ and $Q$ and repeatedly run change detection algorithms
for $100$ times;
then we observe detected edges by CVLL. 
Finally, we select the edges that are detected using the original
non-shuffled dataset and remove those that
were detected in the shuffled datasets for more than $5$ times
(i.e., the significance level $5\%$).
For KLIEP, $k$ is also tuned by using CVLL.
In Figure~\ref{fig.twitter},
we plot detected change graphs which are generated using
samples of $P$ starting from April 17th, July 6th, and July 26th,
respectively.

The initial explosion happened on April 20th, 2010. Both methods discover dependency changes between keywords. Generally speaking, KLIEP captures more conditional independence changes between keywords
than the Flasso method, especially when comparing Figure~\ref{fig.twitter.proposed.july} and Figure~\ref{fig.twitter.fl.july}. 
At the first two stages (\cref{fig.twitter.proposed.april,fig.twitter.fl.april,fig.twitter.proposed.june,fig.twitter.fl.june}),
the keyword ``Obama'' is very well connected with other keywords in the results given by both methods. Indeed, at the early development of this event, he lies in the center of the news stories, and his media exposure peaks after his visit to the Louisiana coast (May 2nd, May 28nd, and June 5th) and his meeting with BP CEO Tony Hayward on June 16th. Notably, both methods highlight the ``gulf-obama-coast'' triangle in \cref{fig.twitter.proposed.april,fig.twitter.fl.april} and the ``bp-obama-hayward'' chain in \cref{fig.twitter.proposed.june,fig.twitter.fl.june}. 

However, there are some important differences worth mentioning. First, the Flasso method misses the ``transocean-hayward-obama'' triangle in \cref{fig.twitter.fl.april,fig.twitter.fl.june}. Transocean is the contracted operator in the Deepwater Horizon platform, where the initial explosion happened.
On \cref{fig.twitter.proposed.july}, the chain ``bp-spill-oil'' may indicate that the phrase ``bp spill'' or ``oil spill'' has been publicly recognized by the Twitter community since then, while the ``hayward-bp-mexico'' triangle, although relatively weak, may link to the event that Hayward stepped down from the CEO position on July 27th.

It is also noted that Flasso cannot find any changed edges in Figure~\ref{fig.twitter.fl.july},
perhaps due to the Gaussian restriction.

\section{Discussion, Conclusion, and Future Works}
\label{conclusion.sec}
In this paper, we proposed a \emph{direct} approach to learning sparse changes in MNs
by density ratio estimation.
Rather than fitting two MNs separately to data and comparing them to detect a change,
we estimated the ratio of the probability densities of two MNs
where changes can be naturally encoded as sparsity patterns in estimated parameters.
This direct modeling allows us to
halve the number of parameters
and approximate the normalization term in the density ratio model
by a sample average without sampling.
We also showed that the number of parameters to be optimized
can be further reduced with the dual formulation,
which is highly useful when the dimensionality is high. 
Through experiments on artificial and real-world datasets,
we demonstrated the usefulness of the proposed method 
over state-of-the-art methods 
including nonparanormal-based methods and sampling-based methods.


Our important future work 
is to theoretically elucidate the advantage of the proposed method,
beyond the Vapnik's principle of solving the target problem directly.
The relation to \emph{score matching} \citep{hyvarinen2005NonNormalized},
which avoids computing the normalization term in density estimation,
is also an interesting issue to be further investigated.
Considering higher-order MN models
such as the \emph{hierarchical log-linear model}
\citep{Schmidt_Convex_Log_Linear}
is a promising direction for extension.

In the context of change detection, we are mainly interested in the situation
where $p$ and $q$ are close to each other
(if $p$ and $q$ are completely different, it is straightforward to detect changes).
When $p$ and $q$ are similar,
density ratio estimation for $p(\boldx)/q(\boldx)$ or $q(\boldx)/p(\boldx)$ 
perform similarly.
However, given the asymmetry of density ratios,
the solutions for $p(\boldx)/q(\boldx)$ or $q(\boldx)/p(\boldx)$ 
are generally different.
The choice of the numerator and denominator in the ratio
is left for future investigation.

Detecting changes in MNs is the main target of this paper.
On the other hand, estimating the difference/divergence between two probability
distributions has been studied
under a more general context in the statistics and machine learning communities
\citep{book:Amari+Nagaoka:2000,eguchi2006interpreting,wang09divergence,sugiyama12bregman,sugiyama13directdiv}.
In fact, the estimation of 
the \emph{Kullback-Leibler divergence} \citep{Annals-Math-Stat:Kullback+Leibler:1951}
is related to the KLIEP-type density ratio estimation method 
\citep{nguyen2010estimating},
and the estimation of the \emph{Pearson divergence} \citep{pearson1900onethe}
is related to the squared-loss density ratio estimation method \citep{kanamori2009leastsquare}.
However, the density ratio based divergences tend to be sensitive to outliers.
To overcome this problem, a divergence measure based on relative density ratios
was introduced, and its direct estimation method was developed \citep{NC:Yamada+etal:2013}.
$L^2$-distance is another popular difference measure between probability density functions.
$L^2$-distance is symmetric,
unlike the Kullback-Leibler divergence and the Pearson divergence,
and its direct estimation method has been investigated recently
\citep{NC:Sugiyama+etal:2013,IEEE-PAMI:Kim+Scott:2010}.

Change detection in time-series is a related topic.
A straightforward approach is to evaluate the difference (dissimilarity)
between two consecutive segments of time-series signals.
Various methods have been developed to identify the difference
by fitting two models to two segments of time-series separately,
e.g., the singular spectrum transform 
\citep{Moskvina03changepoint,Ide07changepoint},
subspace identification \citep{kawahara2007changepoint},
and the method based on the one-class support vector machine
\citep{desobry2005anonline}.
In the same way as the current paper,
directly modeling of the change has also been explored 
for change detection in time-series
\citep{kawahara2012sequential, liu2013changepoint,NC:Sugiyama+etal:2013}.

\section*{Acknowledgements}
SL is supported by the JST PRESTO program and the JSPS fellowship.
JQ is supported by the JST PRESTO program.
MUG is supported by the Finnish Centre-of-Excellence in Computational
Inference Research COIN (251170).
TS is partially supported by MEXT Kakenhi 25730013, and the Aihara Project, 
the FIRST program from JSPS, initiated by CSTP.
MS is supported by the JST CREST program and AOARD.

\appendix
\section*{Appendix: Derivation of the Dual Optimization Problem}
\label{sec.appx.dual}
First, we rewrite the optimization problem \eqref{eq.final.obj} as
\begin{align}
\label{eq.primal.obj.2}
&\min_{\boldtheta, \boldw}  \left[\log\left(\sum_{i=1}^{n_Q}\exp\left(w_i\right)\right) - \boldtheta^\top \boldg + \frac{\lambda_1}{2}\boldtheta^\top\boldtheta + \lambda_2\sum_{u\ge v} \|\boldtheta_{u,v}\| - C\right]\\
&\text{subject to }  \boldw = \boldH^\top\boldtheta, \notag
\end{align}
where
\begin{align*}
\boldw&=(w_1,\ldots,w_{n_Q})^\top,\\ 
\boldH &=
(\boldH_{1,1}^\top,\ldots,\boldH_{d,1}^\top, \boldH_{2,2}^\top,\ldots,\boldH_{d,2}^\top,\ldots,\boldH_{d,d}^\top)^\top,\\
\boldH_{u,v} &=[\boldf(x_1^{(u)Q},x_1^{(v)Q}), \ldots, \boldf(x_{n_Q}^{(u)Q},x_{n_Q}^{(v)Q})],\\
\boldg &=(\boldg_{1,1}^\top,\ldots,\boldg_{d,1}^\top,\boldg_{2,2}^\top,\ldots,\boldg_{d,2}^\top,\ldots,\boldg_{d,d}^\top)^\top,\\
\boldg_{u,v} &=\frac{1}{n_P}\sum_{i=1}^{n_P} \boldf(x_i^{(u)P},x_i^{(v)P}),\\
C &= \log n_Q.
\end{align*}
With Lagrange multipliers $\boldalpha=(\alpha_1,\ldots,\alpha_{n_Q})^\top$,
the Lagrangian of \eqref{eq.primal.obj.2} is given as
\begin{align}
\label{eq.obj.lagrangian}
\mathcal{L}(\boldalpha)& = \min_{\boldw, \boldtheta}
\left[ \log\sum_{i=1}^{n_Q}\exp\left(w_i\right) - \boldtheta^\top \boldg + \frac{\lambda_1}{2}\boldtheta^\top\boldtheta  + \lambda_2\sum_{u\ge v} \|\boldtheta_{u,v}\|
-(\boldw-\boldH^\top\boldtheta)^\top \boldalpha
\right] - C \notag\\
& = \min_{\boldw} \left[\log\sum_{i=1}^{n_Q}\exp\left(w_i\right) -\boldw^\top \boldalpha  \right] \notag\\
&\phantom{=}
+ \min_{\boldtheta} \left[\boldtheta^\top(\boldH\boldalpha-\boldg) + \frac{\lambda_1}{2}\boldtheta^\top\boldtheta + \lambda_2\sum_{u\ge v} \|\boldtheta_{u,v}\|  \right] - C \notag\\
&= \min_{\boldw} \psi_1(\boldw) + \min_{\boldtheta} \psi_2(\boldtheta) - C.
\end{align}

A few lines of algebra can show that $\psi_1(\boldw)$ reaches the minimum
$-\sum_{i=1}^{n_Q} \alpha_i\log \alpha_i$
at
\begin{align*}
\alpha_i = \frac{\exp(w_i)}{\sum_{i=1}^{n_Q} \exp(w_i)}, ~~~ i = 1, \ldots, n_Q.
\end{align*}
Note that extra constraints are implied from the above equation:
\begin{align*}
\alpha_1,\ldots,\alpha_{n_Q}\ge0 \text{ and } \sum_{i=1}^{n_Q}\alpha_i = 1.
\end{align*}

Since $\psi_2(\boldtheta)$ is not differentiable at $\boldtheta_{u,v} = \boldsymbol{0}$, we can only obtain its sub-gradient:
\begin{align*}
\nabla_{\boldtheta_{u,v}} \psi_2(\boldtheta) =
-\boldxi_{u,v}
+ \lambda_1\boldtheta + \lambda_2 \nabla_{\boldtheta_{u,v}}\|\boldtheta_{u,v}\|,
\end{align*}
where 
\begin{align*}
\boldxi_{u,v}&=\boldg_{u,v}-\boldH_{u,v}\boldalpha \notag,\\
\nabla_{\boldtheta_{u,v}}\|\boldtheta_{u,v}\| &=
\begin{cases}
\displaystyle
\frac{\boldtheta_{u,v}}{\|\boldtheta_{u,v}\|} &\text{if } \boldtheta_{u,v} \ne \boldsymbol{0},\\[4mm]
\displaystyle
\{\boldy ~|~ \|\boldy\|\le 1 \}  &\text{if } \boldtheta_{u,v} = \boldsymbol{0}.
\end{cases}
\end{align*}
By setting $\nabla_{\boldtheta_t} \psi_2(\boldtheta) = \boldsymbol{0}$, we can obtain the solution to this minimization problem by Eq.\eqref{eq.dual.solution}.

Substituting the solutions of the above two minimization problems
with respect to $\boldtheta$ and $\boldw$ into \eqref{eq.obj.lagrangian},
we obtain the dual optimization problem \eqref{eq.obj.dual}.

\bibliography{main}

\begin{thebibliography}{48}
\bibitem[Pearson(1900)]{pearson1900onethe}
K.~Pearson.
\newblock On the criterion that a given system of deviations from the probable
  in the case of a correlated system of variables is such that it can be
  reasonably supposed to have arisen from random sampling.
\newblock \emph{Philosophical Magazine}, 50:\penalty0 157--175, 1900.
\end{thebibliography}

\end{document}